\def\year{2020}\relax
\documentclass[letterpaper]{article} % DO NOT CHANGE THIS
\usepackage{aaai20}  % DO NOT CHANGE THIS
\usepackage[square,sort,comma,semicolon,numbers,authoryear]{natbib}
\usepackage{times}  % DO NOT CHANGE THIS
\usepackage{helvet} % DO NOT CHANGE THIS
\usepackage{courier}  % DO NOT CHANGE THIS
\usepackage[hyphens]{url}  % DO NOT CHANGE THIS
\usepackage{graphicx} % DO NOT CHANGE THIS
\usepackage{graphicx}  % DO NOT CHANGE THIS
\usepackage{subfig}
\usepackage{float}
\usepackage{amsmath}
\usepackage{mathbbol}
\usepackage{algorithm}
\usepackage{algorithmic}

\pdfoutput=1
\usepackage{cleveref}
\title{Consensus-based Interpretable Deep Neural Networks with Application to Mortality Prediction}
\author{\Large \textbf{Shaeke Salman,\thanks{Equal contributions}\textsuperscript{\rm 1} Seyedeh Neelufar Payrovnaziri,\footnotemark[1]\textsuperscript{\rm 2} Xiuwen Liu,\textsuperscript{\rm 1}
Pablo Rengifo-Moreno,\textsuperscript{\rm 3,\rm 4}
Zhe He\textsuperscript{\rm 2}}\\ 
\textsuperscript{\rm 1}Department of Computer Science, Florida State University, FL 32306, USA\\ \textsuperscript{\rm 2}School of Information, Florida State University, FL 32306, USA \\
\textsuperscript{\rm 3}College of Medicine, Florida State University, FL 32306, USA \\
\textsuperscript{\rm 4} Tallahassee Memorial Hospital, FL 32308, USA\\
\{salman, liux\}@cs.fsu.edu, spayrovnaziri@fsu.edu,
paren@southern-med.com,
zhe.he@cci.fsu.edu
}
\begin{document}

\maketitle

\begin{abstract}
  Deep neural networks have achieved remarkable success in various challenging tasks. However, the black-box nature of such networks is not acceptable to critical applications, such as healthcare. In particular, the existence of adversarial examples and their overgeneralization to irrelevant, out-of-distribution inputs with high confidence makes it difficult, if not impossible,
to explain decisions by such networks. In this paper, we analyze the underlying mechanism of generalization of deep neural networks and propose an ($n$, $k$) consensus algorithm which is insensitive to adversarial examples and can reliably reject out-of-distribution samples. Furthermore, the consensus algorithm is able to improve classification accuracy by using multiple trained deep neural networks. To handle the complexity of deep neural networks, we cluster linear approximations of individual models and identify highly correlated clusters among different models to capture feature importance robustly, resulting in improved interpretability.  
Motivated by the importance of building accurate and interpretable prediction models for healthcare, our experimental results on an ICU dataset show the effectiveness of our algorithm in enhancing both the prediction accuracy and the interpretability of deep neural network models on one-year patient mortality prediction. In particular, while the proposed method maintains similar interpretability as conventional shallow models such as logistic regression, it improves the prediction accuracy significantly.

\end{abstract}

\section{Introduction}

Cardiovascular diseases (CVDs) cause severe economic and healthcare related burdens not only in the United States but worldwide~\citep{benjamin2018heart}. Acute myocardial infarction (AMI) is a type of CVD which is defined as ``myocardial necrosis in a clinical setting consistent with myocardial ischemia", or heart attack in simple words~\citep{10.1093/eurheartj/ehs215}. AMI is the leading cause of death worldwide~\citep{reed2017acute}. Identifying risky patients in Intensive Care Unit (ICU) and preparing for their health needs are crucial to health planners and health providers~\citep{desalvo2006mortality}. This enables them to appropriately manage AMI and employ timely interventions to reduce mortality from CVD~\citep{mcnamara2016predicting}. These facts motivate recent efforts on building mortality prediction models for ICU patients with AMI ~\citep{article}. Electronic health records (EHRs) with %enormously 
rich data of patient encounters present unprecedented opportunities for critical clinical applications such as 
%phenotyping and  
outcome prediction~\citep{shickel2017deep}. However, medical data are typically heterogeneous and building machine learning models using this type of data is more challenging than using homogeneous data like images. The necessity of dealing with heterogeneous data is not limited to medical applications, but it is shared among many other applications of machine learning~\citep{l2017machine}.

\begin{figure*}[ht]
\centering
%    \captionsetup{justification=centering}
    \includegraphics[scale =0.12]{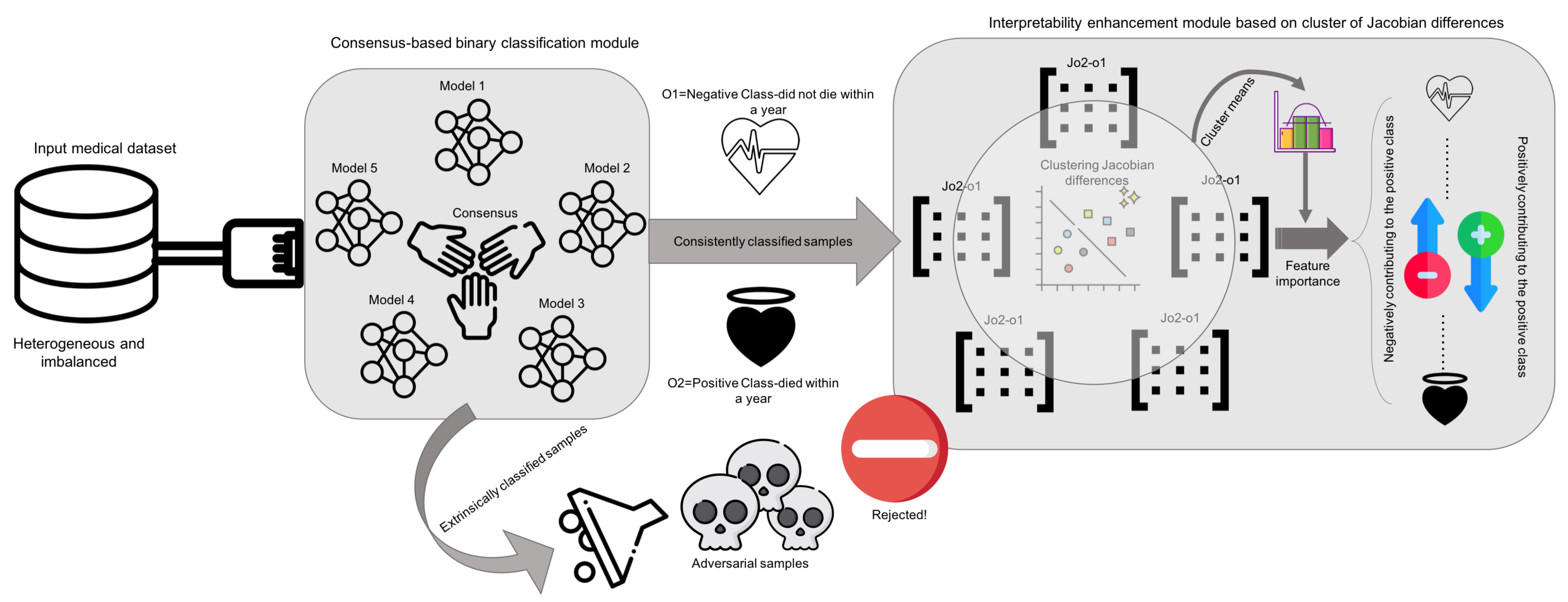}
    \vspace{-0.05in}
    \caption{The workflow of the study (Icons made by https://www.flaticon.com)}
    \label{fig:AAAI-Framework}
\end{figure*}

Unlike traditional machine learning approaches, deep learning methods do not require feature engineering~\citep{steele2018machine}.
%Deep neural 
Such networks have demonstrated significant successes 
%and surpassed human-level performance 
in many challenging tasks and applications~\citep{lecun+al-2015-nature}.
%including computer vision~\cite{AlexNet2012Imagenet,DeepResidual2016He}, machine translation~\cite{SeqToSeq2014Sut}, speech recognition~\cite{SpeechReco2013Alex} and  game playing~\cite{AlphaGo2017Silver2017}. 
Even though they have been employed
in numerous real-world applications to enhance
the user experience, their adoption 
%there is sluggish progress in interpretability enhancement of such successful models. 
in healthcare and clinical practice %such as disease diagnosis, treatment selection, and patient monitoring~\cite{yu2018artificial} 
has been slow. %Despite this valuable progress, 
Among the inherent difficulties, the complexity of these models remains a huge challenge~\citep{ahmad2018interpretable} as it is not clear how they arrive at their predictions~\citep{zhang2018interpretable}. In medical practice, it is unacceptable to only rely on predictions made by a black-box model to guide decision-making for patients. Ideally, medical professionals should be able to point out the conditions under which the model works. Any incorrect prediction such as erroneous diagnosis may lead to serious medical errors, which is currently the third leading cause of death in the United States~\citep{makary2016medical}. This issue has been raised %in the past by several researchers in the related  literature 
and the necessity of interpretable deep learning models has been identified~\citep{adadi2018peeking}. However, it is not clear how to improve the interpretability and at the same
time retain the accuracy of deep neural networks.
Deep neural networks have improved application performance by capturing complex latent relationships among input variables. To make the matter worse, these models are typically overparameterized, i.e., they have more parameters than the number of training samples~\citep{Zhang2016UnderstandingDL}. Overparametrization simplifies the optimization problem for finding good solutions~\citep{GoingBeyondTwo}; however, the resulting solutions are even more complex and more difficult to
interpret.
Consequently, interpretability enhancement techniques would be difficult
without handling the complexity of deep neural networks.

%Researchers introduced deep k-nearest neighbors (DkNN) algorithm to provide confidence, robustness and interpretability to the deep learning model~\cite{papernot2018deep}. In their work, the neighbors serve as the points in the input domain that support the model predictions and are easily interpretable by human. Other researchers proposed a tree-regularization technique which approximates the complex model by a function that is interpretable by human~\cite{wu2018beyond}. They explicitly regularize deep models to make them human-interpretable. This is achieved by closely modeling these complex models using decision trees with few nodes. Others worked on interpretability enhancement by identifying the nonlinear coordinates on which the dynamics are globally linear~\cite{lusch2018deep}. They developed deep neural network representations of Koopman eigenfunctions that are interpretable even in case of high-dimensional and non-linear systems ~\cite{williams2015data}. In another work, researchers showed that training deep neural networks with regularized gradients increases the model interpretability~\cite{ross2018improving}. They also discussed the connection between adversarial robustness and interpretability of the model.

Recognizing that commonly-used activation functions (ReLU, sigmoid, tanh, and so on) are piece-wise linear or can be well approximated by a piece-wise linear function, such neural networks partition the input space into (approximately) linear regions.
%Due to weight symmetry~\cite{ExploringWeight2018Hu}, many of the different linear regions should be equivalent. 
In addition, gradient-based optimization results in similar linear regions for similar inputs as their gradient tends to be similar.
By clustering the linear regions, we can reduce the number of distinctive
linear regions and at the same time improve robustness.
%of the linear approximation. 
To further improve the performance, we train multiple models and use consensus
among the models to reduce their sensitivity to %incidental features 
 adversarial examples with small perturbations and also reduce overgeneralization to irrelevant inputs
of individual models. 
%The resulting algorithm, is called (n, k) consensus algorithm.
%In order to provide linear interpretations of deep neural network models, it is important to understand their underlying mechanism. In this paper, we carefully investigate the behavior of deep neural network models by systematically analyzing adversarial examples~\cite{Intriguing2014Szegedy} as well as overgenaralized examples. Based on our analysis, we proposed (n,k) consensus algorithm which
We demonstrate the effectiveness of deep neural network models and the proposed
algorithms on one-year mortality prediction in patients diagnosed with AMI or post myocardial infarction (PMI) in MIMIC-III database. The workflow of this study is depicted in Fig.~\ref{fig:AAAI-Framework}. Furthermore, the experimental results show that the proposed method improves the performance as well as the interpretability.    

The paper is organized as follows. In the next section, we present generalization and overgeneralization in the context of deep neural networks and the proposed deep ($n$, $k$) consensus-based classification algorithm. %and its effectiveness in handling both generalization and overgeneralization are explained in Section 3. 
After that, we describe a consensus-based interpretability method. Then, we illustrate the effectiveness of the proposed algorithms in enhancing one-year mortality predictions via experiments. Finally, we review recent studies that are closely related to our work and conclude the paper with a brief summary and plan for future work.

\section{Generalization and Overgeneralization in Deep Neural Networks}

Fundamentally, a neural network approximates the underlying but unknown function using $f(x;\theta)$, where $x$ is the input, and $\theta$ is a vector that
includes all the parameters (weights and biases). 
%In a neural network, the functionality of the neurons is decided based on several factors including $f(x;\theta)$, different activation functions (e.g., ReLu, sigmoid, tanh, softmax) and the parameters.
Given a deep learning model and a training dataset, there are two fundamental problems to be solved: optimization and generalization.
The optimization problem deals with finding the parameters $\theta$ by minimizing a loss
function on the training set. Overparametrization~\citep{Neyshabur2018Overpara}
in deep neural networks makes the problem easier to solve by 
increasing the number of good solutions exponentially~\citep{Landscape2017TowardsUG}.

Since there are numerous good solutions, understanding their differences and commonalities is essential to developing more effective multiple-model based methods.
Toward a systematic understanding
of deep neural network models in the input space, one must consider the behavior of these models in case of typical, irrelevant and adversarial inputs (the inputs that are ``computed" intentionally to degrade the system performance)~\citep{goodfellow2014explaining}.   
%[Salman and Dr. Liu, I guess we need to mention that this work has been done in your previous research, the reader suddenly jumps in to images and MNIST, thus the logic is lost]
%To address this concern, we have looked at the adversarial examples, 
% [This para was in IJCAI] As a representative example, we have trained five different deep neural networks on the MNIST dataset. To study systematically how the models respond to irrelevant images, we have used images from the CIFAR10 dataset as they do not contain valid handwritten digits; we have cropped the images and converted them to the same input format of MNIST. Fig. \ref{fig:overgen_adv}(left) shows how the five models agree on irrelevant samples by showing the maximum number of models that agree with each other over the same classification label for samples. It shows that the models respond (almost) randomly to such irrelevant inputs. 
Previously, 
%\citeauthor{salman2019interpretable}~\shortcite{salman2019interpretable} 
we study the response of multiple deep network models to irrelevant images systematically and show that such models behave almost randomly to irrelevant inputs. Furthermore, we study the response of the models on adversarial examples generated by perturbing the inputs with one of the models using fast sign algorithm~\citep{goodfellow2014explaining}. 
%Please see supplementary materials for detailed results.
Please see the previous version of the work for detailed results.

\begin{figure*}[ht]
  \centering
  \includegraphics[width=0.20\textwidth]{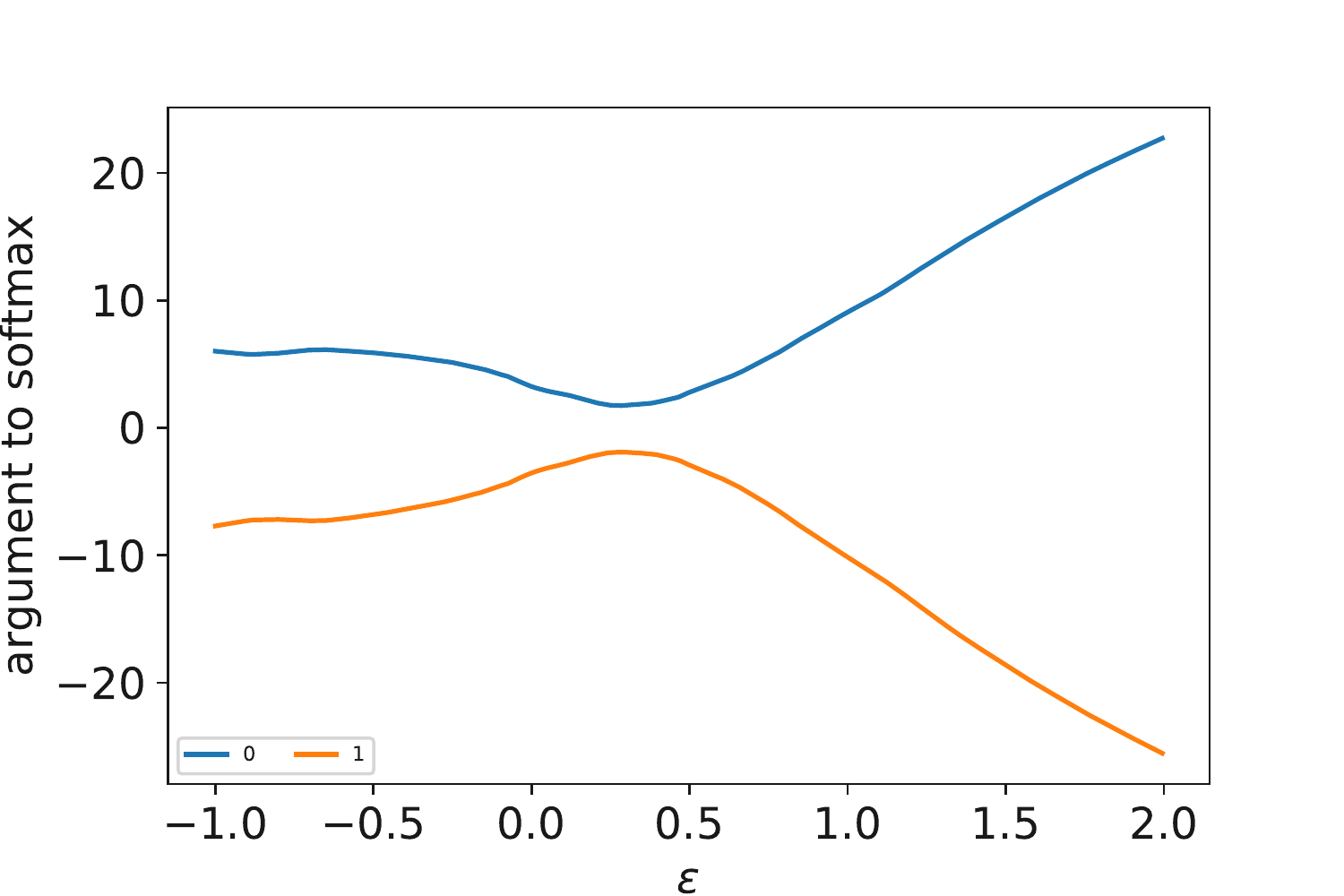}%\label{fig:m1_same_class_0}}
  \includegraphics[width=0.20\textwidth]{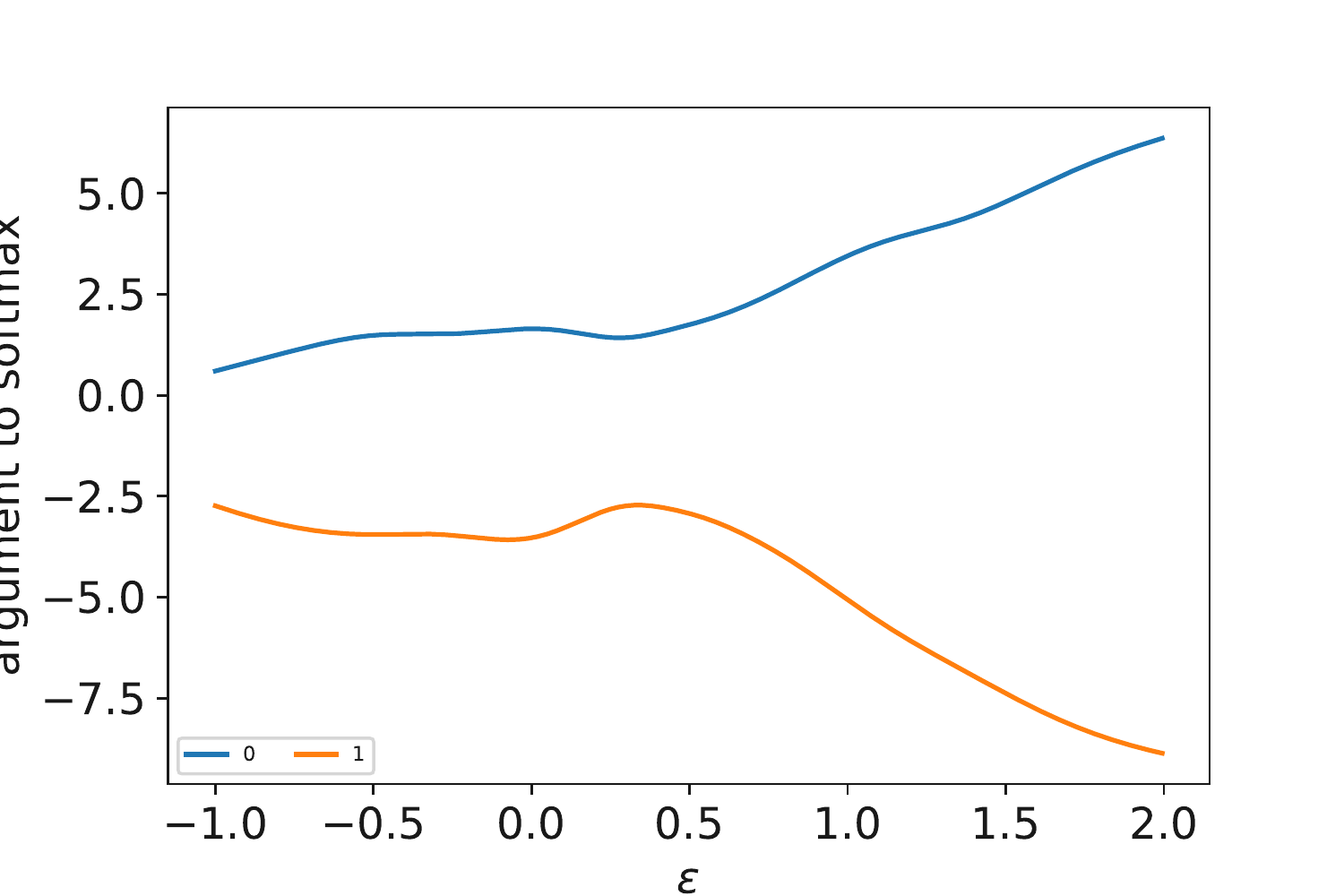}%\label{fig:m2_same_class_0}}
  \includegraphics[width=0.20\textwidth]{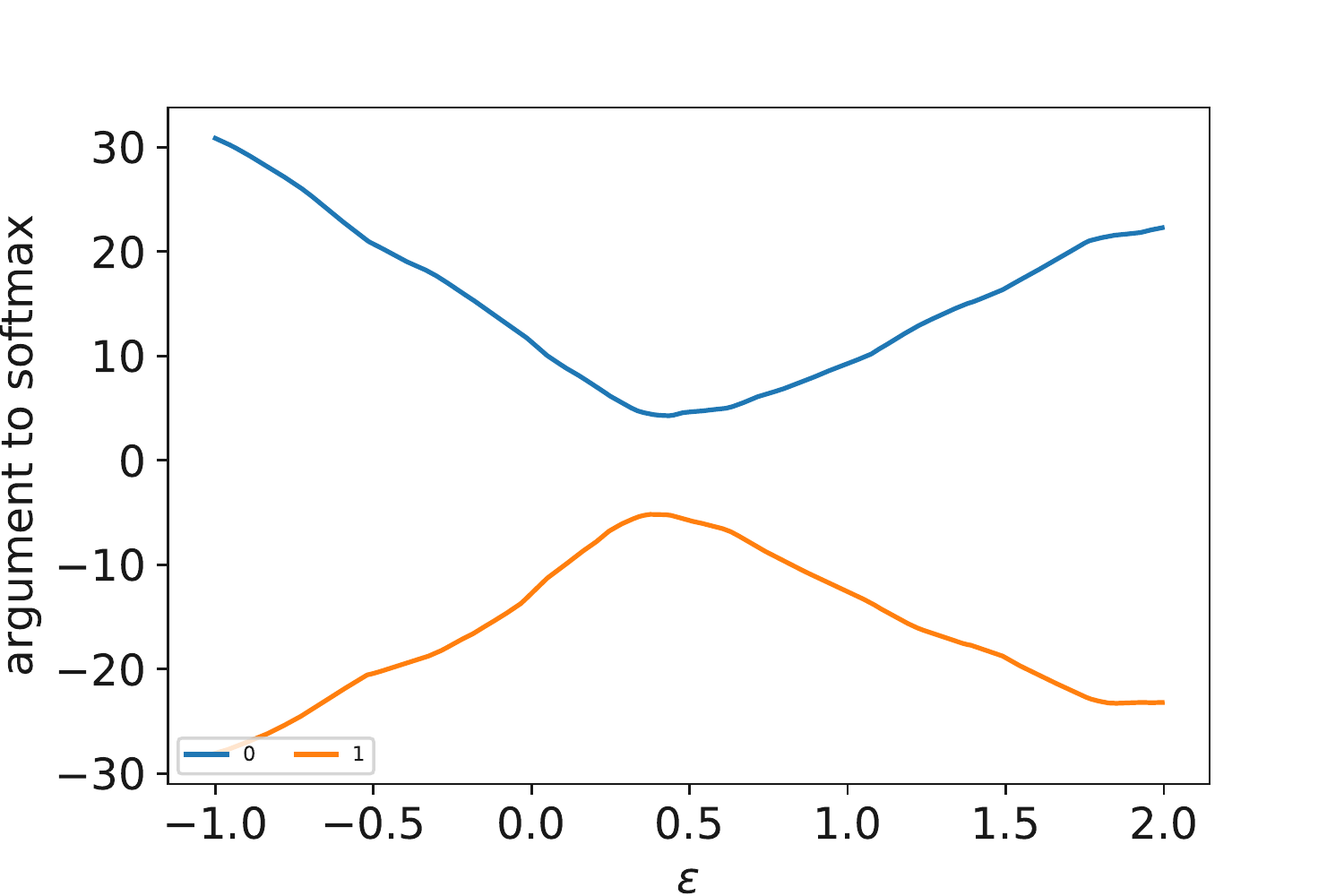}%\label{fig:m3_same_class_0}}
  \includegraphics[width=0.20\textwidth]{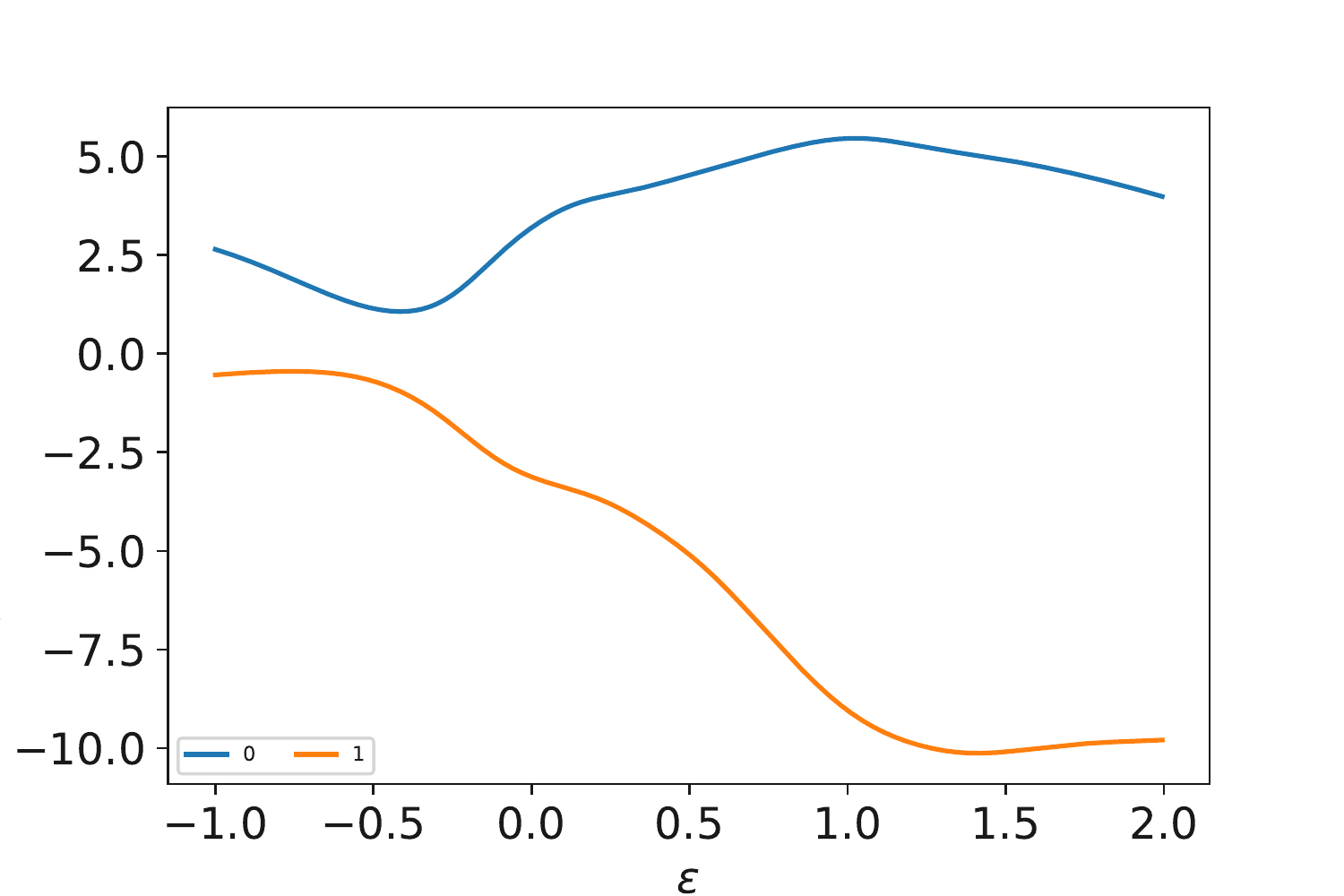}%\label{fig:m4_same_class_0}}
  \includegraphics[width=0.20\textwidth]{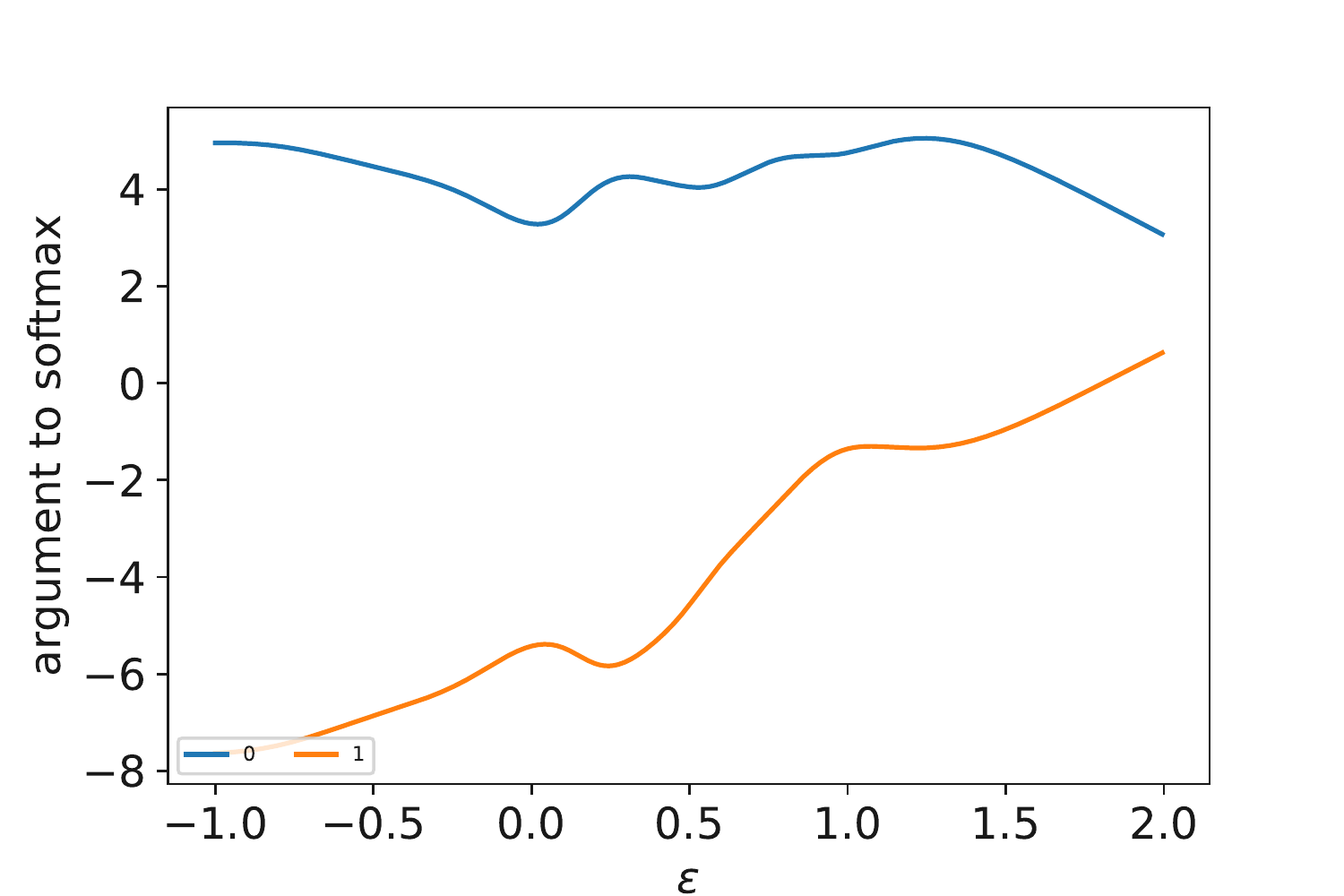}%\label{fig:m5_same_class_0}}
  \vspace{-0.10in}
  \caption{Outputs from the penultimate layer for model 1, 2, 3, 4 and 5 respectively centered at a training sample, along the direction to another sample in the same class.}
  \label{fig:models_same_class_0}
\end{figure*}

\begin{figure*}[ht]
  \centering
  \includegraphics[width=0.20\textwidth]{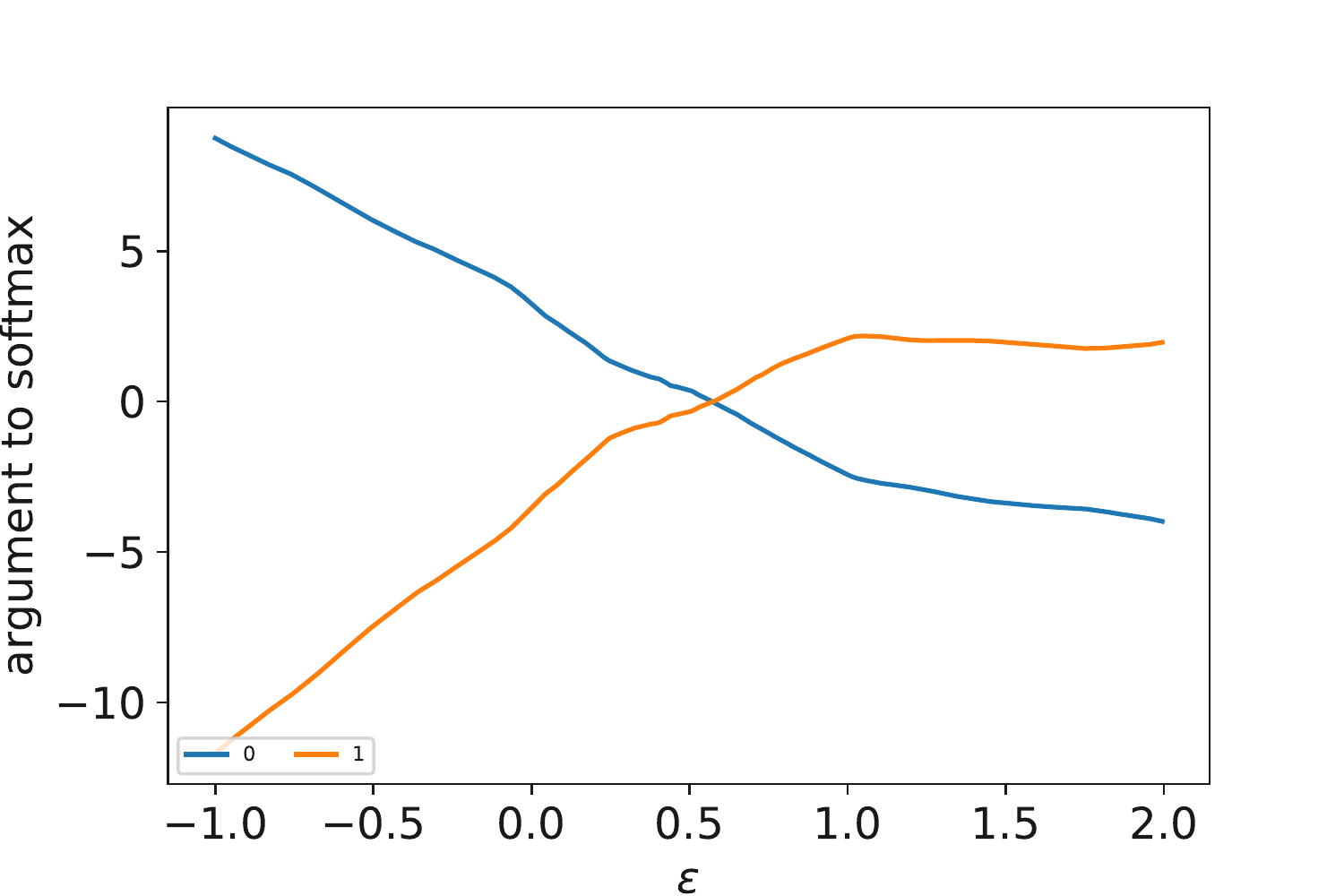}%\label{fig:m1_class_0_1}}
  \includegraphics[width=0.20\textwidth]{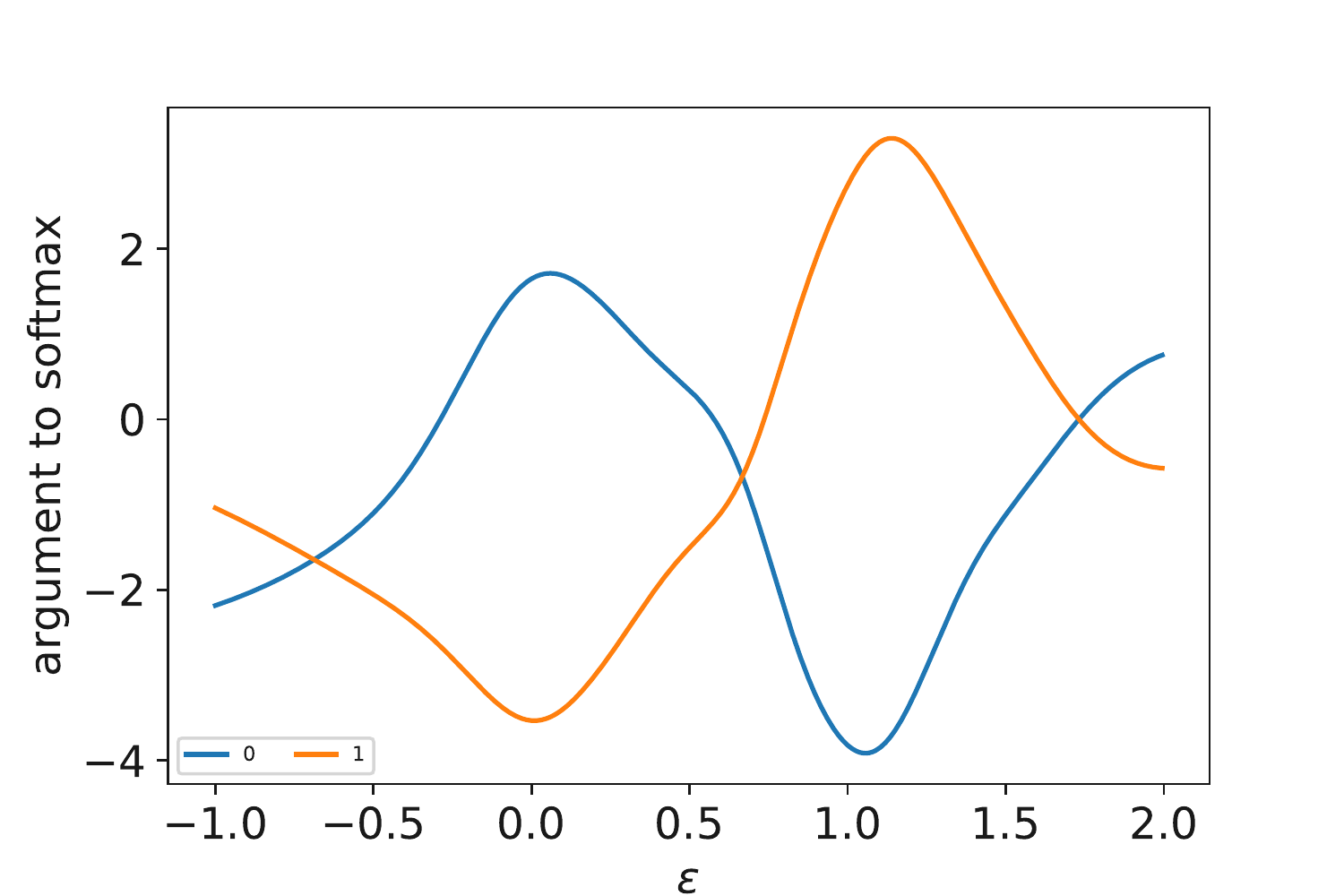}%\label{fig:m2_class_0_1}}
  \includegraphics[width=0.20\textwidth]{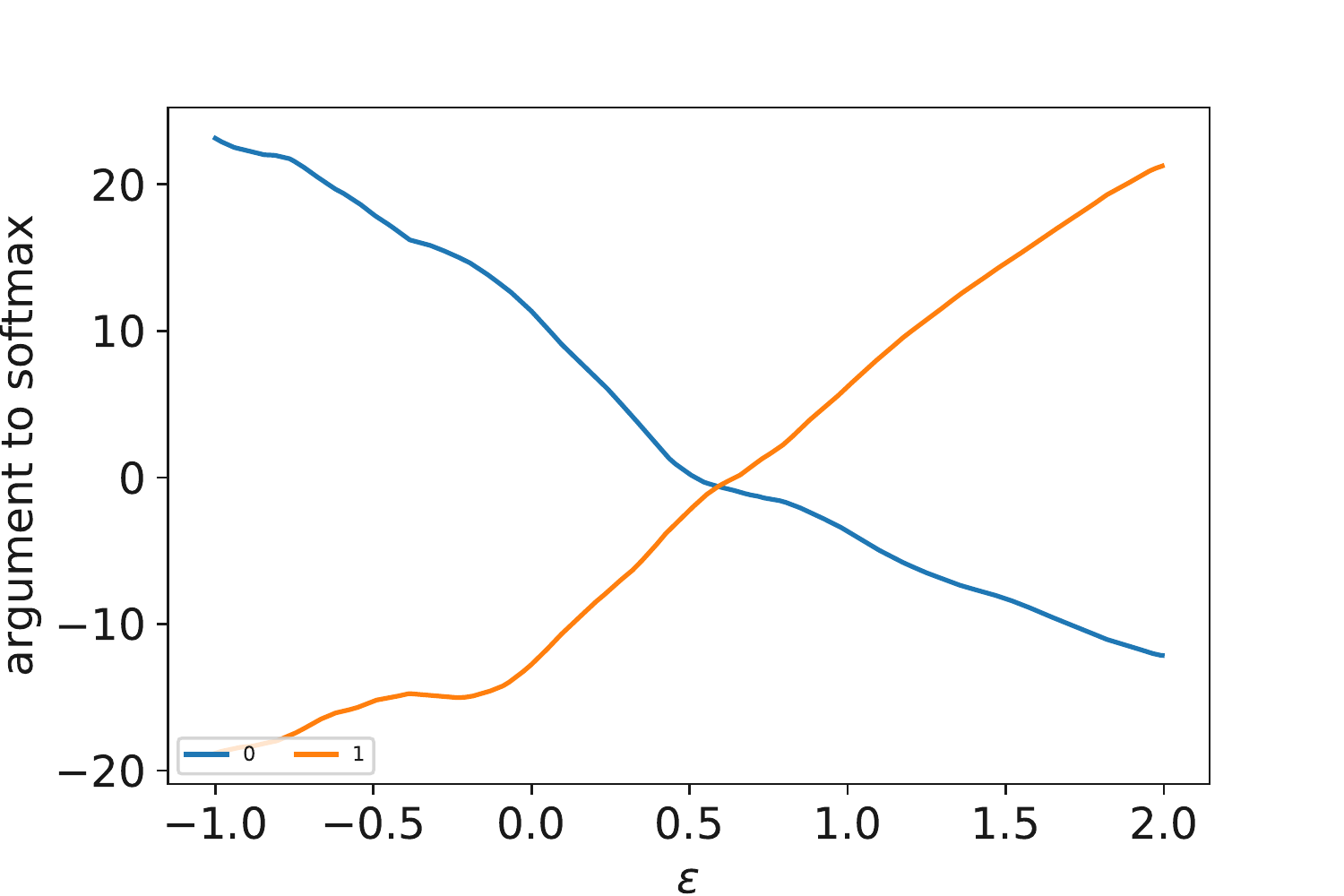}%\label{fig:m3_class_0_1}}
  \includegraphics[width=0.20\textwidth]{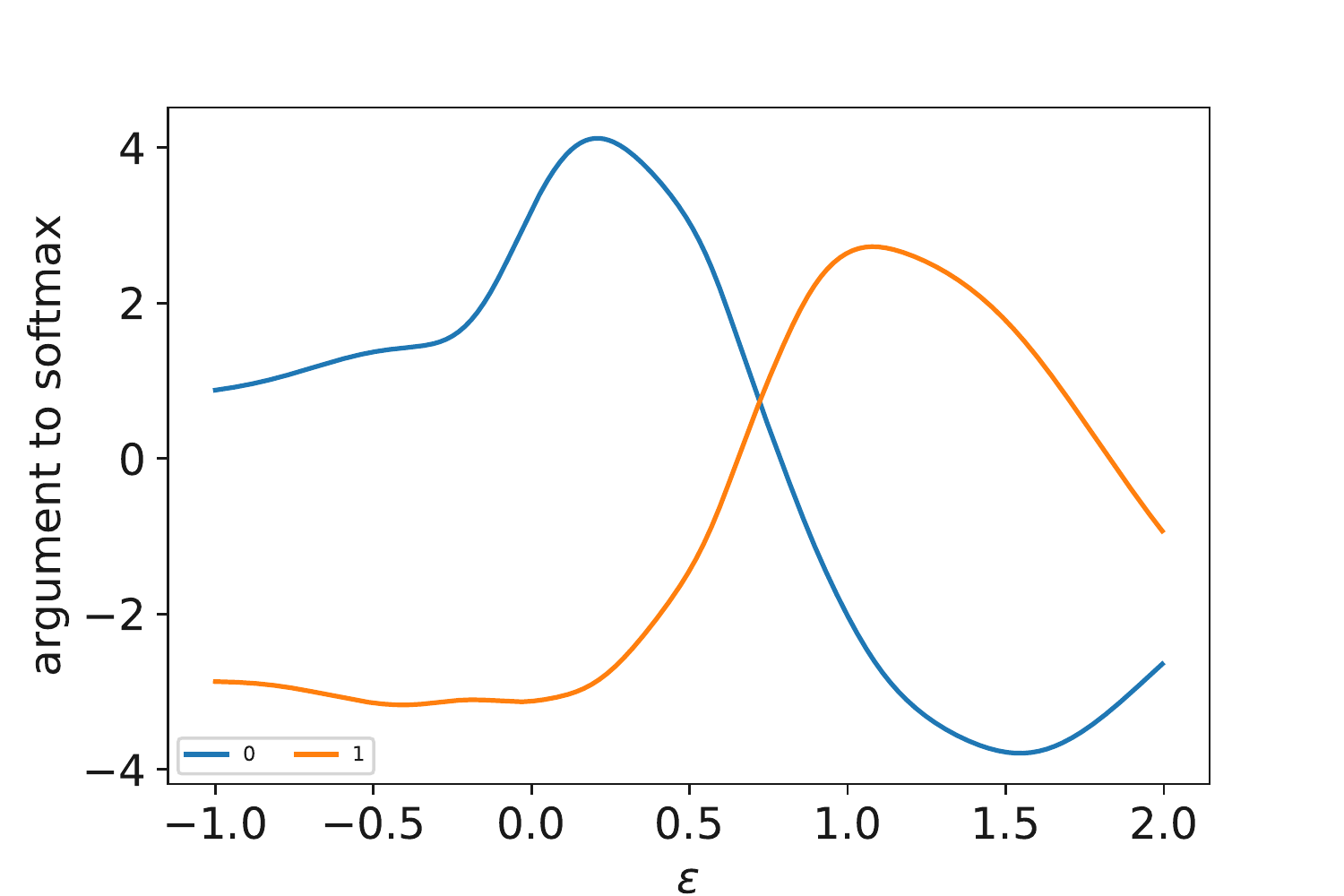}%\label{fig:m4_class_0_1}}
  \includegraphics[width=0.20\textwidth]{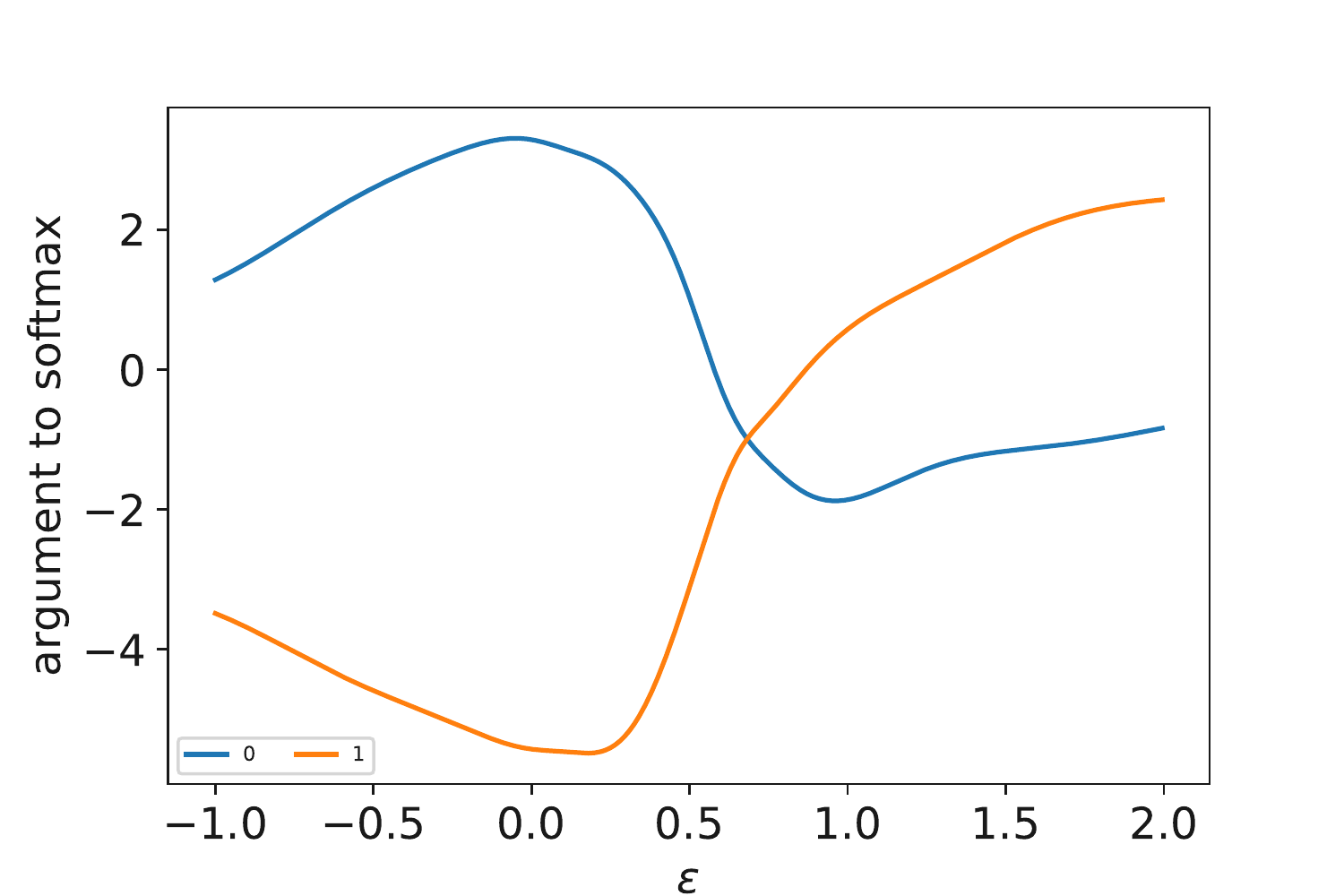}%\label{fig:m5_class_0_1}}
  \vspace{-0.10in}
  \caption{Outputs from the penultimate layer for model 1, 2, 3, 4 and 5 respectively centered at a training sample, along the direction to another sample in the other class.}
  \label{fig:models_class_0_1}
\end{figure*}

\subsection{Deep (n, k) Consensus Algorithm}

As all the models classify training samples accurately, they generate similar linear regions and should 
behave similarly at training samples. 
%While it may appear obvious, improved accuracy is due to a more fundamental reason, where different models generalize well, similarly by learning the (implicit) manifolds in the data (see Fig. 6 (right)). 
In Fig. \ref{fig:models_same_class_0}, the model generalizes perfectly along the path of the same class even though they differ in details. We find this is a representative behavior, main reason why multiple DNN models mostly agree with each other to meaningful inputs and can filter out adversarial and irrelevant samples. While individual approximations are sensitive to adversarial examples, consensus can be used to capture the underlying common structures in training data, not accidental features.
%This intrinsic behavior of deep networks set the basis of our proposed consensus among the models.
%Fig. \ref{fig:models_same_class_0} and Fig. \ref{fig:models_class_0_1} illustrate the basis for consensus. 
%To further illustrate this behavior, we check the generalization of the considered models along a particular direction centered at a training sample. By Fig. \ref{fig:models_same_class_0}, note that they all generalize consistently along the path of the same class even though they differ in details. 
Similar to Fig. \ref{fig:models_same_class_0}, Fig. \ref{fig:models_class_0_1} shows that along the path to other class, all models behave similarly in that the decision boundaries occur around 0.5 and there is no significant oscillation between class 0 and 1 samples. Therefore, consensus between the models makes sense. 

%Further motivated by the overgeneralization and adversarial examples, 
%shown in Fig. \ref{fig:overgen_adv}, 
We propose to use consensus among different models to 
differentiate extrinsically classified samples from intrinsically/consistently classified samples (CCS).
Samples are considered to be consistently classified if they are classified by multiple models with a high probability in the same class.
In contrast, extrinsic factors such as randomness of weight initialization or oversensitiveness to accidental features 
are responsible for the classification of extrinsic samples. As such random factors cannot happen consistently in multiple models, 
we can reduce them exponentially by using more models.
\begin{figure}[ht]
    \centering
    \vspace{-0.15in}
%    \captionsetup{justification=centering}
    \includegraphics[width=7cm]{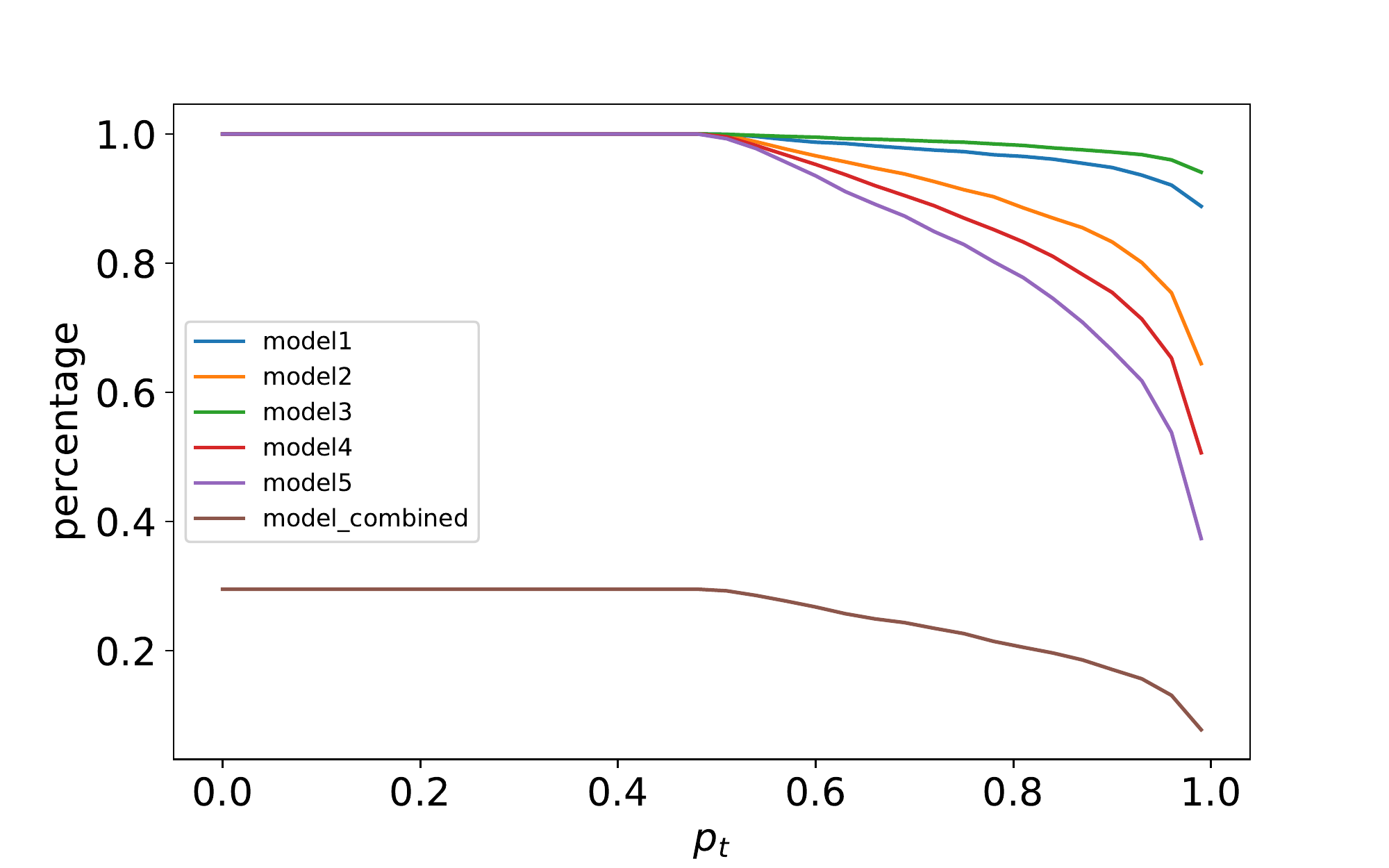}
    \vspace{-0.10in}
    \caption{(to be viewed in color) Percentage of classified overgeneralized samples with (5,5) consensus. The samples are out-of-distribution inputs for the one-year patient mortality prediction dataset.}
    \label{fig:cifar10_ccs_per}
\end{figure}
%To apply our consensus algorithm, we consider two classes of samples - extrinsically classified samples and intrinsically/consistently classified samples (CCS).  In contrast, extrinsic factors such as randomness of weight initialization, and the softmax outputs are responsible for classification of extrinsic samples. We can term those samples as overgeneralized since we think randomness factors cannot happen consistently in multiple models. We have already shown how our algorithm could reduce the effect of the overgenralized samples. This gives us an aspiration to improve the performance on the intrinsic validation samples. We are able to increase the intrinsic accuracy with respect to the threshold parameter $p_t$. 
To tolerate accidental oversensitiveness of a small number of models, 
we propose deep ($n$, $k$) consensus algorithm,
%\footnotemark\footnotetext{While a preliminary version of the algorithm was introduced in~\cite{OverfittingMecha2019Salman}, no justification was provided.}
which is given in Algorithm 1. Note that $P_{min}$ is a vector with one value for each class as it is computed class-wise. 
 Essentially, the algorithm requires consensus among $k$ out of $n$ trained models %$M_1$,\ldots,$M_n$ 
 in order for a sample to be classified; $p_t$, a threshold parameter, is used to decide if the prediction probability of a model is sufficiently high.

\begin{algorithm}
\caption{Deep (n, k) consensus-based classification}\label{algo1}
\begin{algorithmic}[1]
\REQUIRE Trained models $M_1$, $M_2$,\ldots, $M_n$, input $x$, and parameter $p_t$
\STATE Apply each of the models to classify $x$ and retain the probabilities for each class as $P_{M_i}$
\STATE Compute $P_{min}$ by finding the class-wise minimal among top $k$ $P_{M_i}$
\STATE If $max(P_{min}) > p_t$,
\STATE \hspace{0.1in}   Classify $x$ as the class with maximum $max(P_{min})$
\STATE Else 
\STATE \hspace{0.1in}    Reject to classify $x$ (mark it as ambiguous)
\STATE Endif
\end{algorithmic}
\end{algorithm}
%[Dr, Liu, it is our previous algorithm. We need to modify it for (n,k)]

\iffalse
\begin{figure*}[ht]
  \centering
  \subfloat[]{\includegraphics[width=0.25\textwidth]{Images/adversarial_ccs_accuracy_five_models.pdf}\label{fig:adv_acc_5}}
  \subfloat[]{\includegraphics[width=0.25\textwidth]{Images/adversarial_ccs_percentage_five_models.pdf}\label{fig:adv_per_5}}
  \subfloat[]{\includegraphics[width=0.25\textwidth]{Images/adversarial_ccs_accuracy_four_models.pdf}\label{fig:adv_acc_4}}
  \subfloat[]{\includegraphics[width=0.25\textwidth]{Images/adversarial_ccs_percentage_four_models.pdf}\label{fig:adv_per_4}}
  \vspace{-0.10in}
  \caption{Accuracy and percentage when adversarial examples created by model-3 are used in accordance with (5,5) and (5,4) consensus. (a) Accuracy of the models with (5,5) consensus. (b) Percentage of the classified samples with (5,5) consensus. (c) Accuracy of the models with (5,4) consensus. (d) Percentage of the classified samples with (5,4) consensus}
  \label{fig:adv_acc_per}
\end{figure*}
\fi

To illustrate the effectiveness of the proposed algorithm, Fig. \ref{fig:cifar10_ccs_per} %and \ref{fig:adv_acc_per} 
shows
the results on the irrelevant %and adversarial 
samples, generated using randomized values. 
Majority of the irrelevant samples are rejected using a (5, 5) consensus algorithm as shown in Fig. \ref{fig:cifar10_ccs_per};
note that a (5,4) algorithm would also be effective even though it could not reject the ones where four models agree accidentally.

As a whole, the deep network models generalize in a similar way for the intrinsic features. Even though all the models behave consistently for the samples that are supported by the training set, however their behavior can differ in terms of the exact direction that leads to adversarial examples. Therefore, the consensus among the models should also be able to reduce adversarial examples. 
%In a related work, \citeauthor{salman2019interpretable} (\citeyear{salman2019interpretable}) 
We have conducted systematic experiments to illustrate that, where after generating adversarial examples using one model, most of the other models are able to reject them 
(see previous version of the work).
%(see supplementary material).

%[This was in IJCAI one] Fig. \ref{fig:adv_acc_per} shows  that by using a (5,4) consensus algorithm we can classify most of the adversarial examples correctly; the only ones are rejected due to that model m5 misclassified several samples. Clearly, as model m3 is oversensitive to the adversarial examples, a (5, 5) algorithm will reject all the adversarial images.

%these phenomena. This experiment also emphasized the need for (n,k) consensus. Using only (n,n) consensus will cause over-sensitivity. If one of the models is over-sensitive to the samples, then none of the other models will be able to do the classification as well. In case of (5,5) consensus, if the overlap of the prediction scores  is a small value, then the algorithm would not be very useful, because in this case most of the samples will be discarded. As shown in Figure \ref{fig:cifar10_ccs_per},
%by using a (5,5) consensus algorithm, majority of irrelevant samples are 
%rejected. 
Note that the proposed algorithm is different from ensemble methods~\citep{Ensemble2017Ju}, which are used to improve the performance of multiple
models via voting.
%For example, ensemble methods will not be able to handle most of the irrelevant samples while our algorithm is very ffective in rejecting irrelevant samples (as shown in Fig. \ref{fig:cifar10_ccs_per}).
%Ensemble methods do not necessarily reduce the overgeneralization, while they might negatively impact the accuracy of the model. On the other hand, our proposed consensus model relies on the fact that different deep neural network models tend to agree on the prediction results with each other. In consensus-based models, one concern is empty intersection of the models (contradiction). 
Similarly, in applications (such as data mining) where a set of samples need to be classified by possibly multiple classifiers at the same time, correlations between different classes can be utilized to create multiple labels for similar objects by maximizing agreements among the assigned labels to the objects (e.g., \cite{Xie2013MultiConsensus}). Note that these methods can not recognize and reject adversarial and out-of-distribution samples while our algorithm is designed to handle those samples via inherent consensus of the multiple deep neural network models. In addition, the proposed consensus-based algorithm is not trying to force or maximize the consensus among the models for the classification task, rather because the consensus exists naturally in deep networks for the samples they can generalize.

%\section{A New Interpretability Method}
\section{A Consensus-based Interpretability Method}

With a robust way to handle irrelevant and adversarial inputs,
we propose a novel method to interpret decisions by trained
deep neural networks, based on that such networks behave linearly locally and linear regions form clusters due to weight symmetry. 

The linear approximation, 
%while simple, 
reveals rich deep neural network model behavior in the neighborhood of a sample.
Interesting characteristics of the model can be uncovered by walking along
certain directions from the sample. 
%[Next line was written on mnist case. But with binary case, do we need this line?]
For example,  adversarial examples are clearly evident along the direction shown in Fig.
%\ref{fig:arg_to_softmax}(left),
\ref{fig:models_class_0_1},
where the classification changes
quickly outside $\epsilon = 0$ (where the given training sample is).
On the other hand, 
%Fig. \ref{fig:arg_to_softmax}(right)
Fig. \ref{fig:models_same_class_0} 
shows robust classification along this particular direction.    
%Figure \ref{fig:piecewise_plane} and Figure \ref{fig:arg_to_softmax} illustrate this concept.
%modified from ijcai one based on new figures with 157 features 
\begin{figure*}[ht]
  \centering
  \subfloat[]{\includegraphics[width=0.25\textwidth]{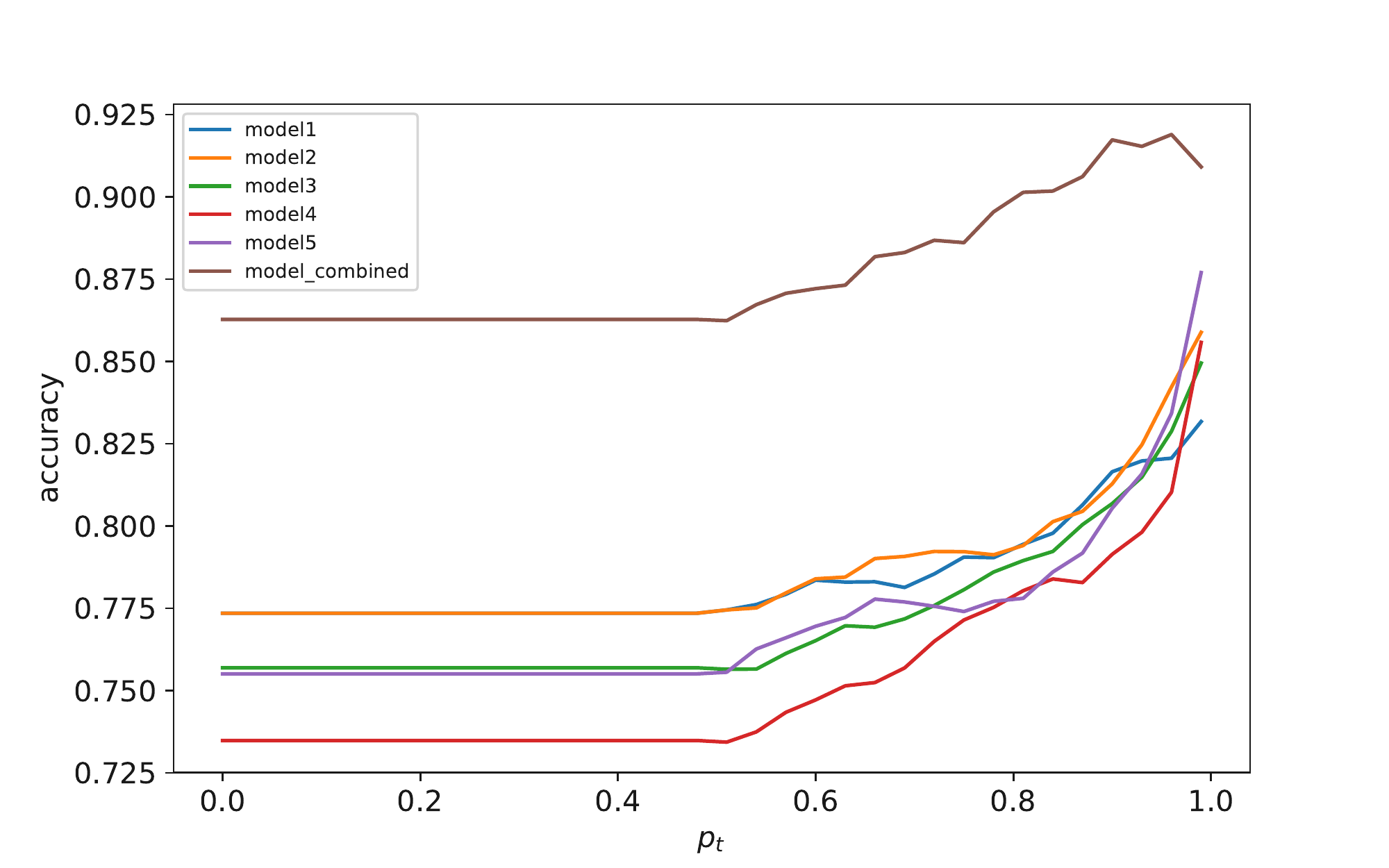}\label{fig:55accm}}
  \subfloat[]{\includegraphics[width=0.25\textwidth]{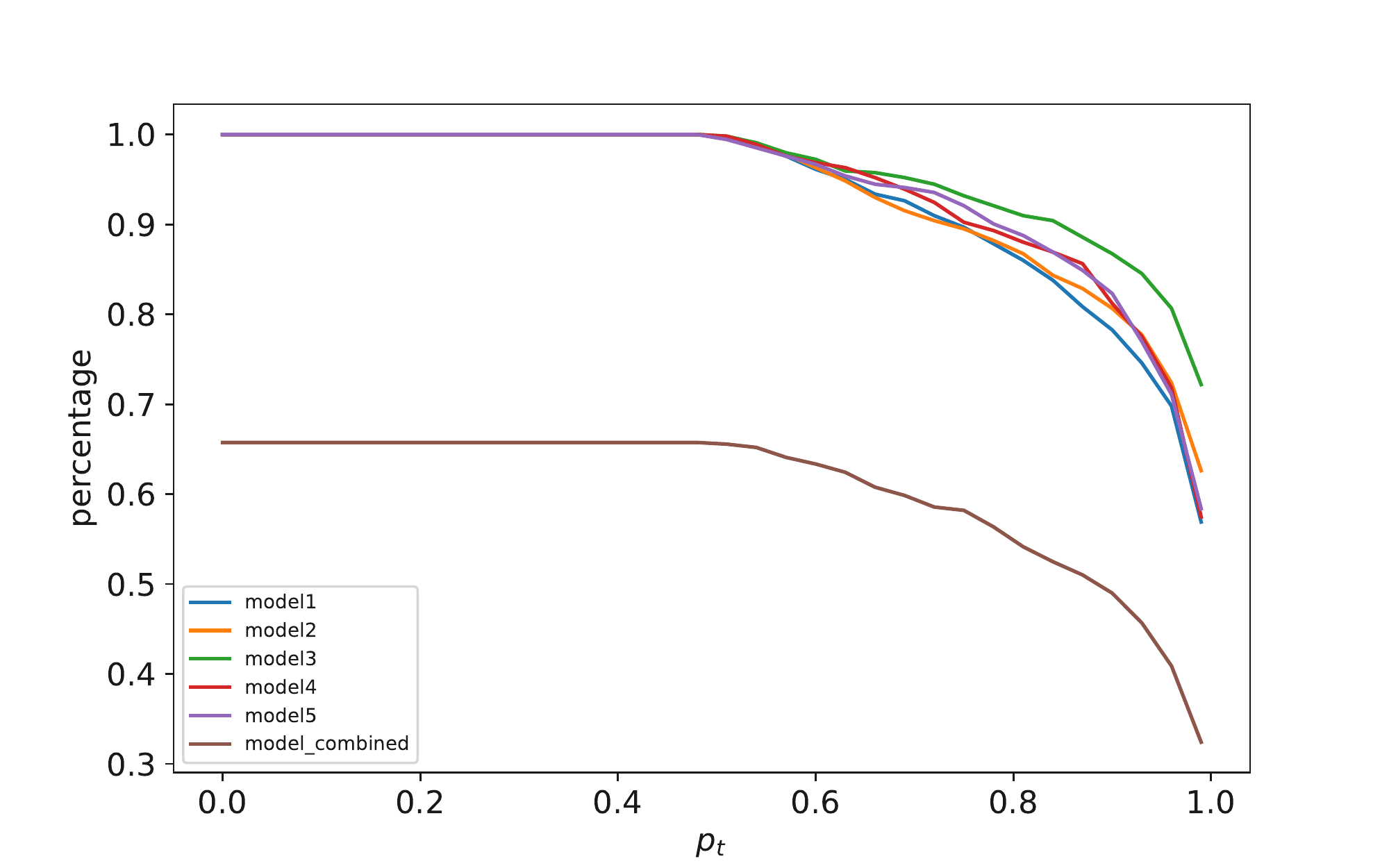}\label{fig:55perm}}
  \subfloat[]{\includegraphics[width=0.25\textwidth]{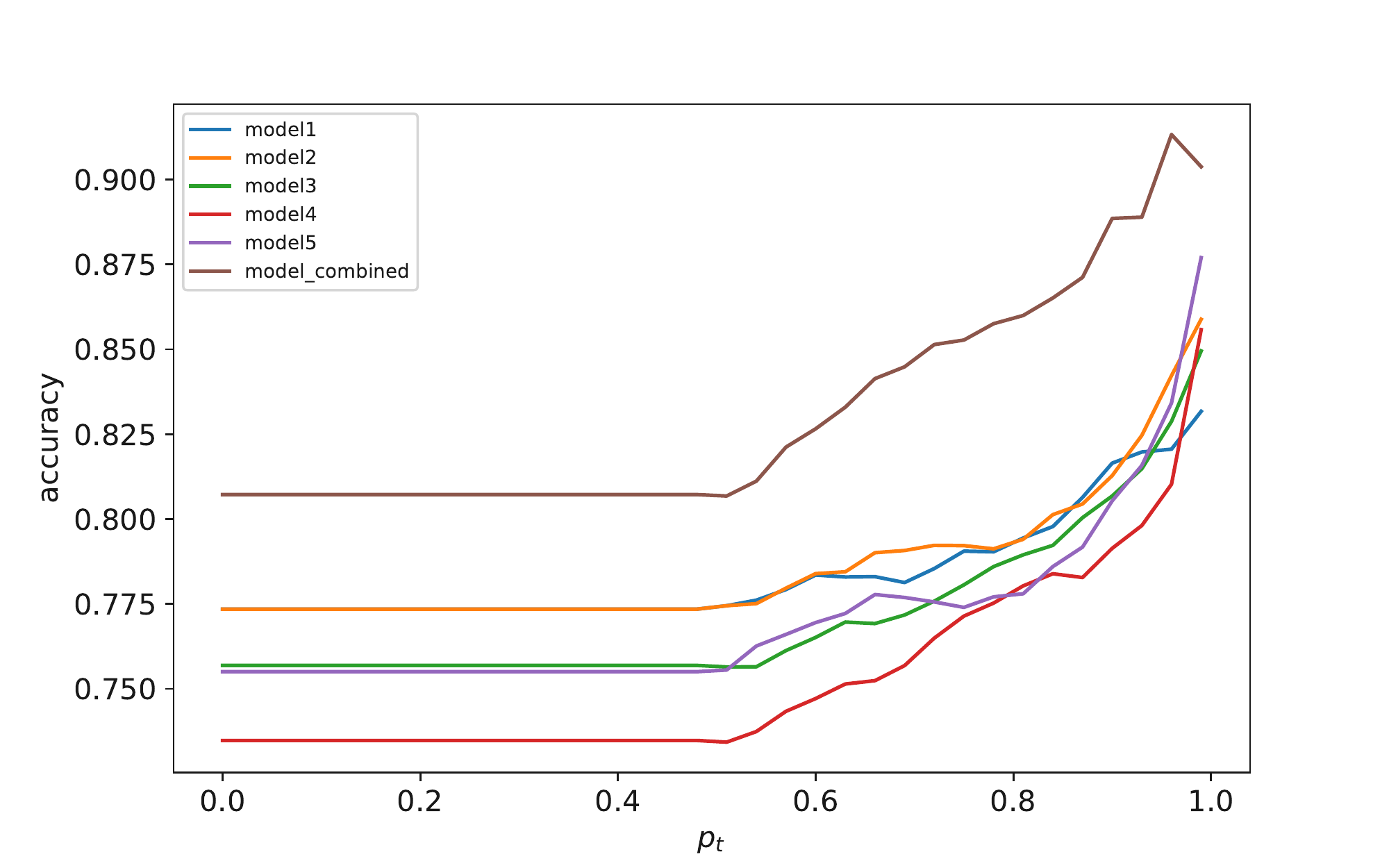}\label{fig:54accm}}
  \subfloat[]{\includegraphics[width=0.25\textwidth]{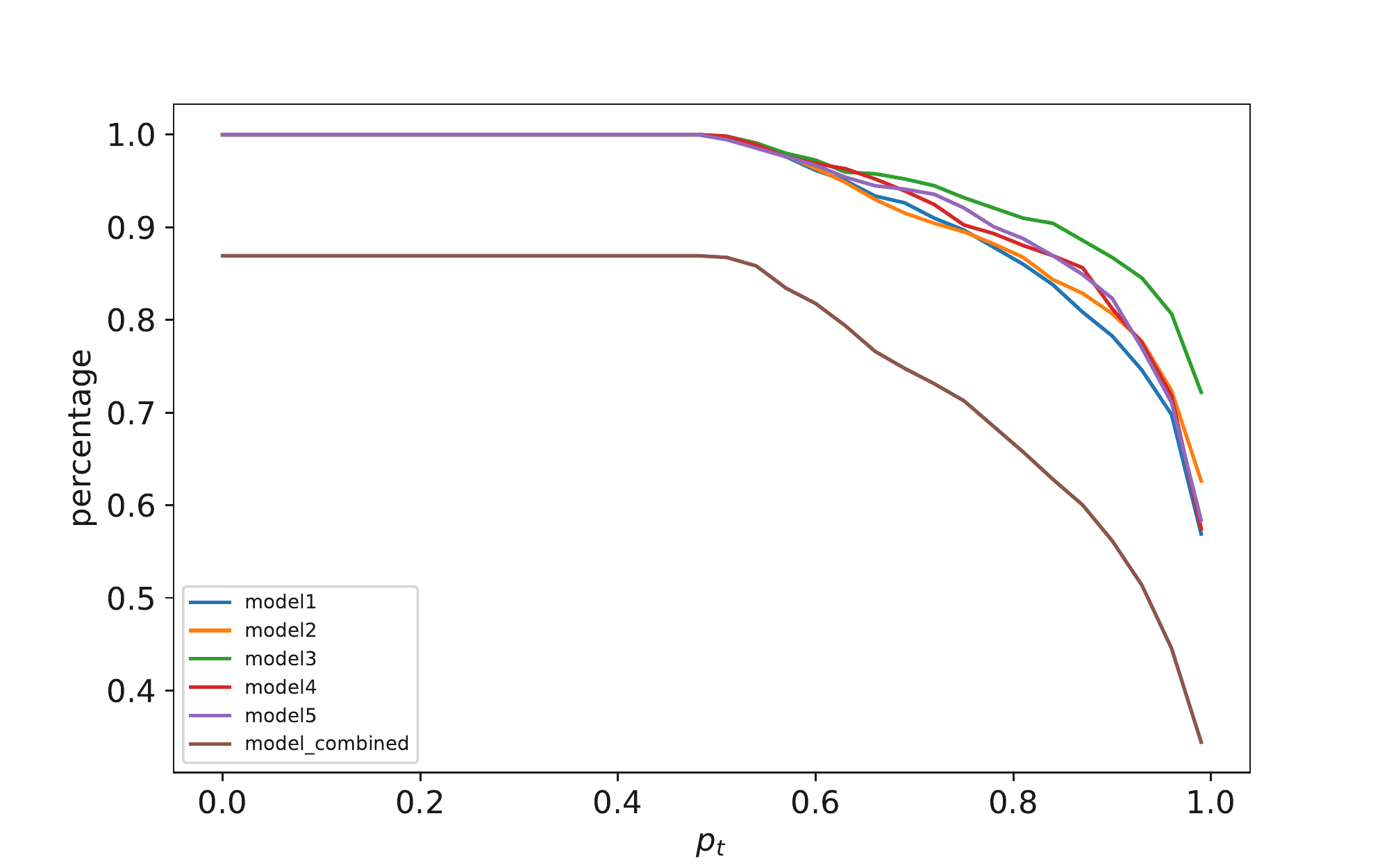}\label{fig:54perm}}
  \vspace{-0.10in}
  \caption{(to be viewed in color) Change of accuracy and percentage of CCS samples with deep ($n$, $k$) consensus on the one-year patient mortality prediction dataset. (a) shows the increase of intrinsic accuracy while (5,5) consensus. (b) shows the percentage of CCS samples while (5,5) consensus. (c) shows the increase of intrinsic accuracy while (5,4) consensus. (d) shows the percentage of CCS samples while (5,4) consensus.}
  \label{fig:55_54_acc_per_mortality}
\end{figure*}

More formally, under the assumption that the last layer in
a neural network is a softmax
layer, we can analyze the outputs from the penultimate layer (i.e., the layer before the softmax layer). Using the notations introduced earlier, the outputs can be written as the following:
\begin{equation} \label{eqn1}
\mathbf{O} = f(\mathbf{x},\theta),
\end{equation}
where $\mathbf{O}$ is the vector-valued function. 
%The behavior of a deep neural network model is determined by this function. 
%In this setting, the resulting decision regions (the inputs that are classified to the same output class) are piece-wise linear. 
Since the model is locally close to linear, if we perturb an input (e.g., $\mathbf{x}_0$) by a small value $\Delta x$ (i.e., $\mathbf{x} = \mathbf{x}_0+\Delta x$), then the Equation \ref{eqn1} can be approximated using the first order Taylor expansion 
around input $\mathbf{x}_0$.
\begin{equation} \label{eqn2}
\mathbf{O} \approx f(\mathbf{x}_0,\theta) + \mathbf{J}\Delta x
\end{equation}
Here, $\mathbf{J}$ is the Jacobian matrix of function $\mathbf{O}$, defined as $J_{i,j} = \frac{\partial O_i}{\partial x_j}$. 
%The Jacobian matrix contains all the partial derivatives of the function $\mathbf{O}$.
Note that the gradient or the Jacobian matrix, in general, has been used
in a number of methods to enhance interpretability (e.g., \cite{simonyan2013deep,Axiomatic2017Sundararajan}).

However, the Jacobian matrix only reflects the changes for each output individually. As classification is inherently a discriminative task, the difference between the two largest outputs is locally important.
In the binary case, we can write the difference of the two outputs as:
\begin{equation} \label{eqn3}
o_2 - o_1 = f_2(\mathbf{x}_0,\theta) - f_1
(\mathbf{x}_0,\theta) + (\mathbf{J}_{1,:}-\mathbf{J}_{0,:})\Delta x,
\end{equation}
where $\mathbf{J}_{0,:}$ and $\mathbf{J}_{1,:}$ are the first and second row of $\mathbf{J}$. In general, for multiple class cases, we need to analyze the difference between the top two %or top $k$ 
outputs locally. For example, we can focus on the difference between the top 2 classes, even though there are 10 classes in case of MNIST dataset~\citep{Mnist1998Lecun}. In general, we can apply pairwise difference analysis; however, most pairs will be irrelevant locally. %Again Fig. 6, 
The differences between bottom classes are likely due to noisy.

The Jacobian difference vector essentially determines the contributions of changes in the features, i.e., the feature importance locally. This allows us to explain why the deep neural network model behaves in a particular way in the neighborhood of a sample. Note that the first part of Equation \ref{eqn3}, i.e., $f_2(\mathbf{x}_0,\theta) - f_1(\mathbf{x}_0,\theta)$ is important to achieve high accuracy. 
%This can be linearly approximated locally.
%Sensitivity analysis is to analyze these linear approximations further. The model interpretation with regards to an individual sample might be noisy. 
However, the local Jacobian matrices, while important, are not robust.
%In addition, many of the local Jacobian matrices behave similarlydue to weight symmetry~\cite{ExploringWeight2018Hu}. 
To increase the robustness of interpretation and at the same time 
reduce the complexity, we propose to cluster the difference vectors
of Jacobian matrices. 
%The number of clusters reflects the model complexity (which is determined by the data complexity). Studying the clusters allows us to quantitatively understand  how a trained deep neural network model generalizes to new samples in a computationally efficient and effective way.
%So we believe the clustering will enhance the interpretability and also increase the robustness of the interpretability. 
%Figure \ref{fig:density_avg_jacobian}(b) shows the average of the local difference vector (given in Equation \ref{eqn3}) and it reveals a robust pattern of significant features. 

The Jacobian difference vectors can be clustered using K-means 
or any other clustering algorithm. In this paper,
we identify consistent clusters using the correlation coefficients of the Jacobian difference vectors of the training samples. To create a cluster, we first identify the pair that has the highest correlation. Then, we expand the cluster by adding the sample with the highest correlation with all the samples in the cluster already. This can be done efficiently by computing the minimum correlations to the ones in the cluster already
for each remaining sample and then choosing the one with the maximum. %Thus, for each sample the minimum correlations with all the other samples %in the set is determined. Then, by taking the maximal of the minimum correlations we choose the potential sample for the cluster. This process guarantees that whatever sample we assign to a potential cluster has high correlations with all the other samples that are already assigned to that cluster. 
We add samples iteratively until the maximum correlation is below a certain threshold. To avoid small clusters, we also impose a minimum cluster size. 
We repeat the clustering process to identify more clusters.
Due to the equivalence of local linear models, the number
of clusters is expected to be small. 
Our experimental results support this.
%; see Section 4.4 for examples.
Note that neural networks still have different biases at different samples, enabling them to classify samples with high accuracy with 
a small number of linear models. 
%[This line was in IJCAI one] This could explain the apparent deep neural network paradox~\cite{Zhang2016UnderstandingDL}: while deep neural networks have many parameters, they typically generalize well also, contradicting traditionally statistical learning theory~\cite{Stat1998Vapnik}. We will investigate this further systematically. 
%. We can plot the distribution of the coefficients (e.g., Figure \ref{fig:density_avg_jacobian} (a)). This gives us idea about the structure of the cluster. The peak of the distribution indicates the presence of either a big cluster or multiple clusters. For each cluster, we can calculate the average which can represent the interpretation for the cluster. Since we make sure the correlations are high, the interpretation should be consistent. After that, we identify the group of clusters that have large correlation coefficients among the cluster averages. In this case, after taking average on a group of clusters, the features with larger Jacobian difference values have more influence on the outcome of the model.   

\iffalse
This is why We cluster the similar samples with k-means algorithm. We set k=5/7/9. We would plot two figures of gradient difference for two of these clusters. Those values should be sorted to get positive and negative influence of the features. [This should be in experimental results section].
\fi

We do clustering for each of the models first. 
The clusters from different models can support each other with strong
correlations between their means and can also complement each other by capturing different aspects of the data. Therefore, we group highly correlated clusters to get more robust interpretations. Note that different subsets of clusters have different interpretations based on the correlations among the clusters.

Given a new sample (e.g., validation sample), we need to check if that sample can be classified correctly by the models at first. If the sample is rejected by the deep $(n, k)$ consensus algorithm, we do not interpret such sample for which models are not confident. In contrast, if it can be classified, we estimate the Jacobian difference for each of the models and then compare that with the cluster means to identify the clusters that provide the strongest support. Note that cluster means are given by consensus of multiple models. 
%see whether that interpretation is similar or not with the clusters. In this process, 
This allows us to check that the new sample is not only classified correctly but also its interpretation is consistent with the interpretation for training samples. 
With multiple models with clusters, the interpretation is  more 
robust.

\begin {table*}[!ht]
\centering
\begin{tabular}{||c|c|c|c|c|c||}
 \hline 
 Specification & Model 1 & Model 2 & Model 3 &Model 4& Model 5\\
 \hline \hline
 Neurons& 200 & 200 &250 &250 & 300 \\
 \hline
 Activation& relu & tanh & relu & tanh & tanh \\
 \hline
 Optimization& SGD & Adamax &Adadelta & Adamax & Adagrad\\
 \hline
 Bias & zeros&ones &costant &ones &random normal\\
 \hline
 Weights &random uniform &random uniform &random normal
 &random uniform & random normal\\
 \hline
\end{tabular}
\caption{Implementation detail of five individual models}
\label{table:1}
\end {table*}

%--need to add some sentence to connect consensus with interpretability

\section{Experimental Results on One-year Mortality Prediction}

\subsection{Dataset}
The Medical Information Mart for Intensive Care III (MIMIC-III) database is a large database of de-identified and comprehensive clinical data which is publicly available. This database includes fine-grained clinical data of more than forty thousand patients who stayed in critical care units of the Beth Israel Deaconess Medical Center between 2001 and 2012. It contains data that are associated with 53,432 admissions for patients aged 16 years or above in the critical care units~\citep{johnson2016mimic}. 

In this study, only those admissions with International Classification of Diseases, Ninth Revision (ICD-9) code of 410.0-411.0 (AMI, PMI) or 412.0 (old myocardial infarction) are considered. These criteria return 5436 records. We use structured data to train the deep neural network models. 
%This is proved to enhance the model performance~\cite{wang2016deep}. 
Structured data includes admission-level information about admission, demographic, treatment, laboratory and chart values, and comorbidities. More details on the features used in this work can be found in a recent work~\citep{barrett2019building}.

\begin {table}[!ht]
{\small
\vspace{-0.075in}
\centering
\begin{tabular}{||c|c|c|c|c|c||}
 \hline 
 {\tiny Model} & {\tiny Accuracy} & {\tiny ROC} & {\tiny Precision} & {\tiny Recall} & {\tiny F-measure}\\
 \hline \hline
 1& 0.7421 & 0.6906 & 0.5849 &0.5568 & 0.5705 \\
 \hline
 2& 0.7679 & 0.7259 & 0.6242 & 0.6167 & 0.6204 \\
 \hline
 3& 0.7348 & 0.6953 & 0.5657 & 0.5928 & 0.5789\\
 \hline
 4 &  0.7513 & 0.7006 &0.6012 &0.5846 & 0.5846 \\
 \hline
5 & 0.7495 &  0.6993 & 0.5974
 &0.5688 & 0.5828\\
 \hline
\end{tabular}
\caption{Evaluation result of five individual models}
\label{table:2}
}
\vspace{-0.225in}
\end {table}

\begin {table*}[!ht]
\centering
\begin{tabular}{|c|c|c|c|c|c|c|}
 \hline 
 Feature & Level of Contribution & Stand dev&	Mean & Median & Minimum&Maximum \\
 \hline \hline
sodium& 46.54&3.32&138.63&138.73&97&160.27\\
 \hline
glucose	& 19.74 &43.6023&	141.4247&	130.47&	51	&543\\
\hline
%admitAge & 18.05 & 13.35& 70.62& 72.21&	18.70& 100.76\\
\hline
%potassium & 17.62& 0.36 &	4.18&	4.14 &1.6 &6.9 \\
%\hline 
%heartRate&	14.02&	12.34&	82.34&	82.34&	26.35	&139.48\\
%\hline
%calcium&	-4.26&	0.57&	8.49&	8.49&	2.6&	13.95\\
%\hline
%systolic&	-14.27&	22.05& 106.67&	106.67&	19.6 &484.12 \\
%\hline
%hemoglobin&	-17.24& 1.53&	10.96&	10.68&	4.31&	18.7\\
%\hline
bicarbonate	&-25.60&	3.61&	24.82&	25&	7 &47.57\\
\hline
chloride &-28.29&	4.29&	103.81&	103.91&	80.42&	125.61\\
 \hline %\\[-0.15in]
 %\hline %[-0.15in]
% \hline
%\vspace{-0.15in}
\end{tabular}
\vspace{-0.075in}
\caption{Some examples of positively and negatively  contributing lab-test features to the ``died within a year" class.}
\label{table:3}
\vspace{-0.125in}
\end {table*}

\begin {table}[!ht]
{\small
\centering
\begin{tabular}{||c|c|c|c|c|c||}
 \hline 
 {\tiny Model} & {\tiny Accuracy} & {\tiny ROC} & {\tiny Precision} & {\tiny Recall} & {\tiny F-measure}\\
 \hline \hline
 LR& 0.7845 & 0.7162 & 0.6923 &0.5389 & 0.6060 \\
 \hline
 SVM& 0.7826 & 0.7066 & 0.7024 & 0.5089 & 0.5902\\
 \hline
 CSVM& 0.7794 & 0.7257 & 0.6577& 0.5868 & 0.6202 \\
 \hline
 Consensus&  0.8623 & 0.87 & 0.7631& 0.6516 & 0.7030 \\
 \hline

 \hline
\end{tabular}
\caption{Comparing the performance of LR, SVM, and CSVM with Consesnsus-based model.}
\label{table:4}
\vspace{-0.225in}
}
\end {table}

\subsection{Results from Individual Models}

Five different deep neural network models are trained for the purpose of this work. Each of these models consists of three dense layers and a softmax layer for classification. 
%These models are different from each other in terms of their number of neurons, activation functions, optimization techniques, and bias and weights initialization. 
%For the implementation purpose, we use the Keras library running on
%top of the Tensorflow framework, as well as a number of
%other Python packages including SciPy, Scikit-learn,
%NumPy, and Pandas.
%[Neelufar, just trying to get some space. If we have space we can add these citations that are commented out]
%For implementation purposes we used the Keras library~\cite{chollet2015keras} running on
%top of the Tensorflow framework~\cite{tensorflow2015-whitepaper}, as well as a number of
%other Python packages including SciPy~\cite{scipy}, Scikit-learn~\cite{scikit-learn},
%NumPy~\cite{numpy}, and Pandas~\cite{mckinneypandas}.
%A detailed information on five individual deep models' implementation is provided in the Supplementary Material. 
Table \ref{table:1} provides implementation details of these models. 

The five models are trained using the same 90\% of the records in the dataset that were randomly selected and evaluated on the remaining 10\%. 
%Same training and evaluation sets are used for all five models. 
%The dataset has 30\% positive and and 70\% negative cases. 
%We decided to leave it as is to reflect this typical challenge with real-world data and investigate if our algorithm could handle it. 
All the values are normalized to between 0 and 1. The evaluation results of the five models are provided in Table \ref{table:2}. The overall accuracy, while varying from model
to model, is in general agreement with other methods.

\subsection{Results from the deep (n, k) Consensus Algorithm}

Here we illustrate the results using the proposed deep ($n$, $k$) consensus algorithm.
%~\cite{OverfittingMecha2019Salman}. 
Fig. \ref{fig:55_54_acc_per_mortality} illustrates its effectiveness
%of the (n,k) consensus algorithm 
on one-year mortality prediction task. It depicts the comparison between the results from individual models and the consensus of the models. Fig. \ref{fig:55_54_acc_per_mortality}(a) and \ref{fig:55_54_acc_per_mortality}(b) show that when the threshold is low (e.g., $p_t<0.5$), (5,5) consensus achieves around 86\% accuracy which is substantially higher than any single model, with around 67\% of the test samples classified. 
%In the (5,5) consensus case, the amount of discarded samples is relatively high for this dataset. 
We also check the effect of the (5,4) and (5,3) versions on the same dataset and observe that (5,4) consensus (i.e., Fig. \ref{fig:55_54_acc_per_mortality}(c) and \ref{fig:55_54_acc_per_mortality}(d)) works well also for this one-year mortality prediction dataset. %Applying (5,4) consensus, 
For  $p_t<0.5$, it provides around 81\% accuracy with around 88\% of the test samples classified. In all the ($n$, $k$) cases, we observe that the number of correctly classified samples among all the consistently classified ones increases with the threshold.

For DNN models, as they generalize similarly, the result is not sensitive to the choice to k (in most cases, n-1 or n should work well). 
%Note that adversarial examples rarely happen in real-world samples; in other words, the situations in Fig. 4(a) and (b), while illustrating an important point of the proposed algorithm, would not occur on real datasets with noticeable probability.  
In general, for DNNs, k should be close to n and $p_t$ should be 0.5 or higher. Different values of the parameters do allow one to fine-tune the trade-off between accuracy and percentage of classified samples. One can choose these two parameters based on how much one would like to emphasize more: accuracy or robustness. 
%For example, with k= 5, with $p_t$ around 0.5, we got about $86\%$ accuracy with $67\%$ percentage, while with k= 4 ($p_t$ same), the numbers are $81\%$ and $87\%$. 
What percentage of samples should be retained, that should be as large as possible, while the accuracy should be as large as possible.

\subsection{Interpretability Models}

To systematically examine the proposed method, we first compute the Jacobian of the training samples and then compute the pairwise correlations. %Then, We convert them into vectors using element-wise differences. 
%[This para was in IJCAI one] %Fig. \ref{fig:density_avg_jacobian} shows an example. The mean correlation among the 11,963,386 pairs is 0.876 with a standard deviation value of 0.035. This deep neural network model has yielded highly consistent and robust interpretation for all the training samples. Note that the biases are different for training samples, enabling high classification accuracy.
\iffalse
\begin{figure}[ht]
%  \centering
  \vspace{-0.16in}
  \captionsetup{width=0.48\textwidth}
  \hspace{-0.12in}
  \includegraphics[width=0.258\textwidth]{Images/density_plot_correlations_model1_std.pdf}%\label{fig:avg_density}\hspace{-0.20in}
%  \hspace{-0.20in}
  \includegraphics[width=0.258\textwidth]{Images/first_model_average_of_jacobian_difference.pdf}%\label{fig:avg_jacobian}}
  \vspace{-0.10in}
  \caption{Left: Distributions of the pairwise correlation coefficients between the difference of the Jacobian rows for a binary classification problem. Right: Average of the Jacobian difference vector.}
  \label{fig:density_avg_jacobian}
%  \vspace{-0.15in}
\end{figure}
\fi
%[replaced the cluster avg figs of ijcai one with new figs with 157 features]
\begin{figure}[ht]
%  \centering
%\vspace{-0.15in}
  \captionsetup{width=0.48\textwidth}
  \hspace{-0.12in}
  \includegraphics[width=0.258\textwidth]{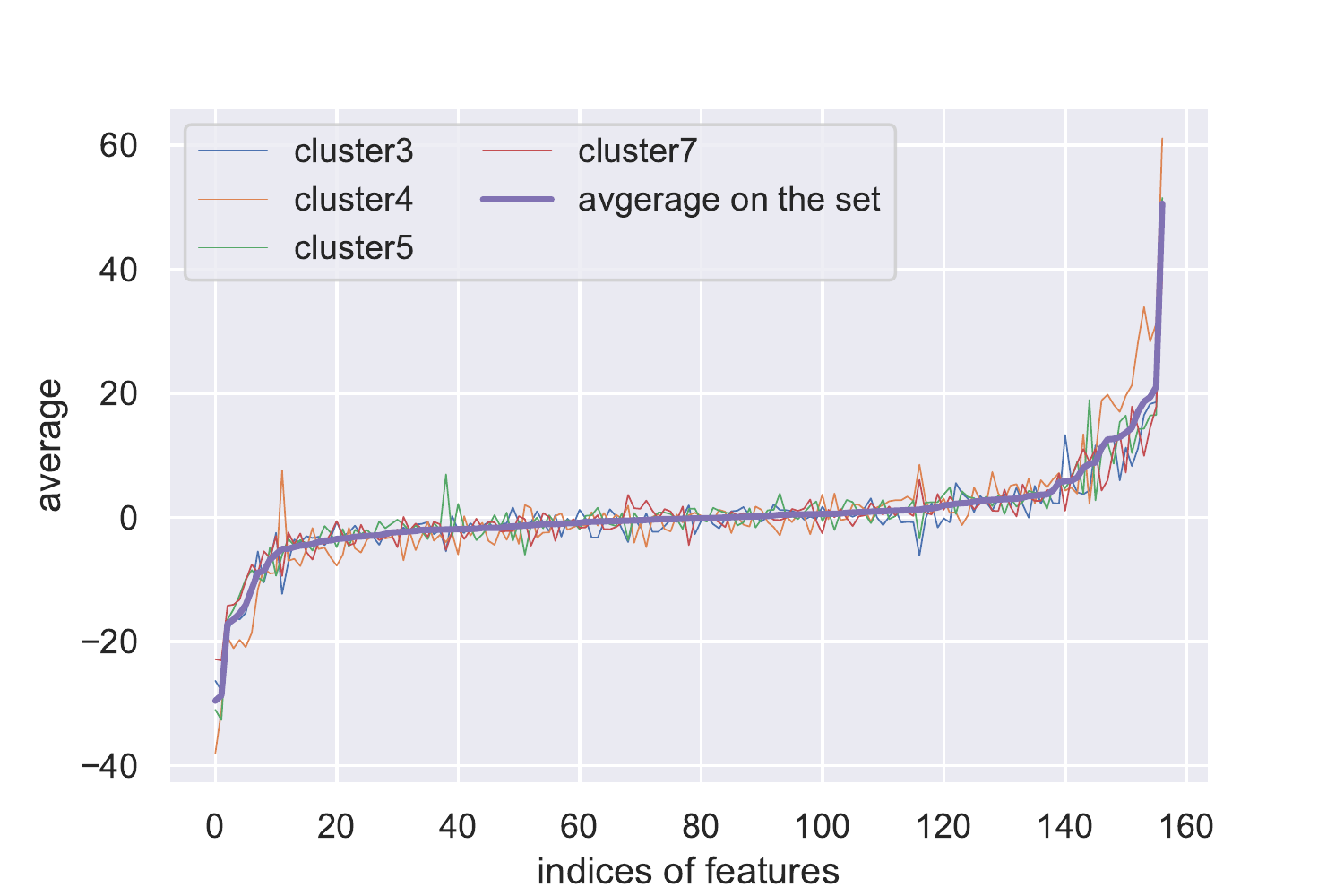}%\label{fig:1_2_7_10_avg_jacobian}}
  \includegraphics[width=0.258\textwidth]{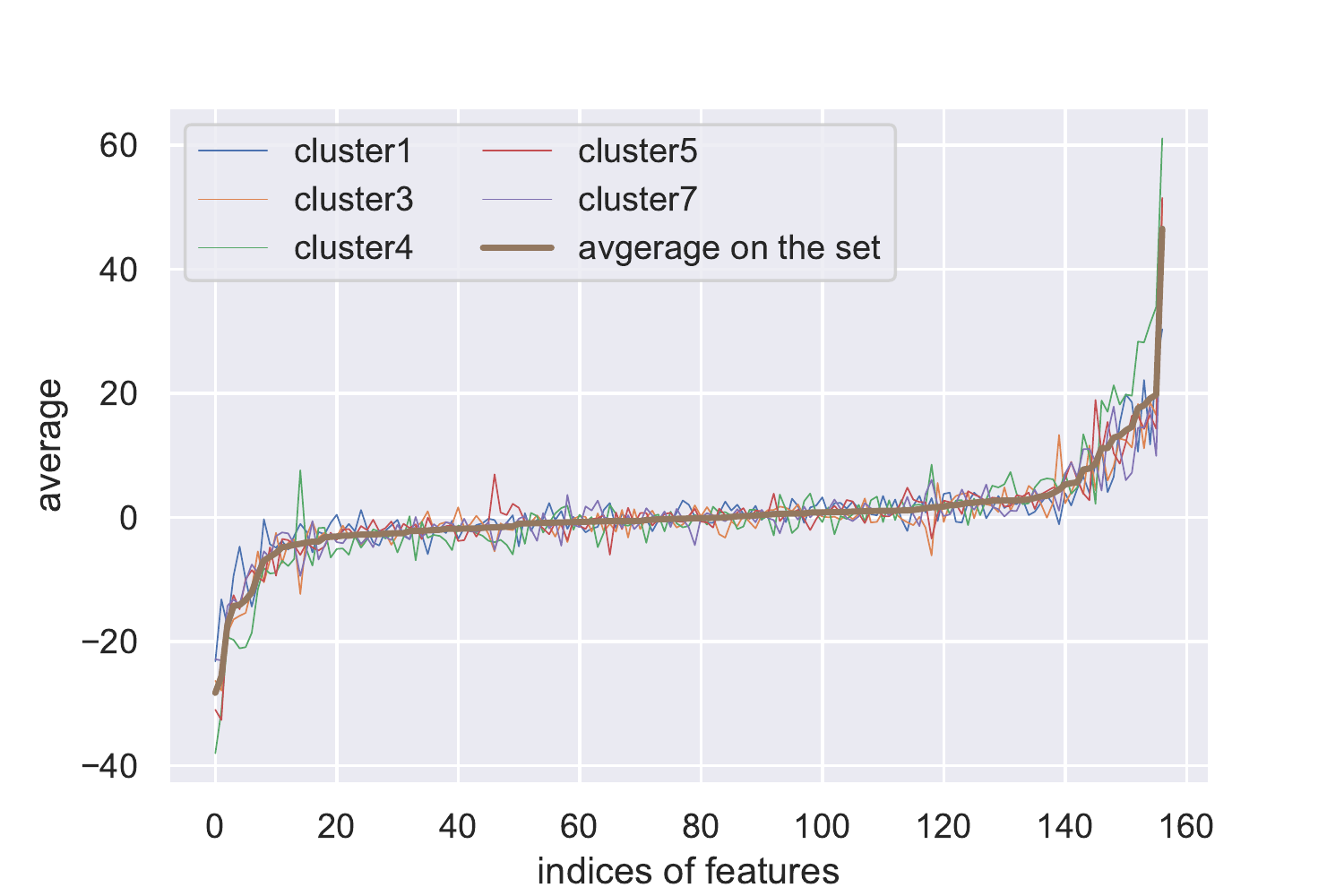}%\label{fig:2_1_3_7_10_avg_jacobian}}
  \vspace{-0.10in}
  \caption{(to be viewed in color) Average of the Jacobian difference vector of highly correlated cluster set. The thicker smooth curve depicts the average on the whole set. The other curves show the average on each cluster of that particular set. Left: First cluster subset. Right: Second cluster subset}
  \label{fig:two_cluster_set_avg_jacobian}
\end{figure}
%[If the interpretation makes sense by Dr. Zhe and Neelufar]
%Since the correlations are consistently high, we calculate the average of the Jacobian difference vectors, which is shown in 
%As we have 369 features (with indices 0 to 368), the calculated average consists of 369 values. We sort those values and plot them in
%Fig. \ref{fig:density_avg_jacobian}(right). It shows that the higher values (i.e., extreme values - leftmost negative or rightmost positive ones) of the average vector correspond to the most relevant and important features. 
As described in the section illustrating the consensus-based interpretabilty method, we group highly correlated clusters to achieve  more robust interpretations. %With this goal
On this dataset, we have considered two subsets of highly correlated clusters among 8 representative clusters by our clustering algorithm.
%illustrated in section 4 (e.g., model 1 with 1 cluster, model 2 with 2 clusters, model 3 with 2 clusters, model 4 with 3 clusters, model 5 with 2 clusters). 
Fig. \ref{fig:two_cluster_set_avg_jacobian} depicts the averages of these subsets along with individual cluster averages. The higher values (i.e., extreme values - leftmost negative or rightmost positive ones) of the average vector correspond to the most relevant and important features. 
%These values are sorted to illustrate negative to positive influence of features. Moreover, each  is also shown accordingly.
%to the feature indices of the sorted average on the cluster set. 
Since they are highly correlated, we notice similar behavior to the average on the subset for each of the clusters. Based on the sorted average of the first subset, we observe that leftmost features in the list have negative impact and rightmost features have positive impact on the positive class (``died within a year"). For the second subset, we notice almost identical features with the positive and negative impact on the positive class. To interpret validation samples, we look at the correlations with each subset.

As a result, we have found that a specific set of features contributes positively to the ``died within a year" class while some other set of features contributes positively to the ``did not die within a year" class. Also, some features show neutral behavior to the classification task, which are placed in the middle of the spectrum with slight tendencies towards either positive or negative ends of the spectrum. Due to space limitations, we illustrate the contributions of only selected features. We have excluded ethnicity and religion-related features since most of them show neutral effect on the prediction outcome. Table \ref{table:3} shows some examples of the most positive, negative features contributing to the positive class. 

Note that the proposed algorithm is inherently scalable. 
Computing Jacobian is done implicitly by backpropagation; as such, the Jacobian
difference vectors can be computed similarly to training the models for one epoch
 using mini batches.
 Furthermore, datasets of billion vectors can 
 be clustered efficiently using product and residual vector quantization techniques (e.g.,~\cite{Matsui2017BillionClust}).

\begin {table*}[h!]
\centering
\begin{tabular}{|c|c|c|c|c|c|c|c|c|c|c|}
 \hline
 n-features & & & & & & & & & & \\
 (top n/2 pos, top n/2 neg)&
 10&
 20&
 30&
 40&
 50&
 60&
 70&
 80&
 90&
 100\\
 \hline \hline
SVM& 
0.8&
0.7&
0.8&
0.725&
0.68&
0.683&
0.728&
0.762&
0.777&
0.8
\\
 \hline

 LR& 0.9&
0.7&
0.766&
0.675&
0.62&
0.683&
0.7&
0.762&
0.766&
0.78
 \\
 \hline

\end{tabular}
\caption{Comparison of the similarity of resulting feature-importance list from intrinsically interpretable models and proposed consensus-based deep model.}
\label{table:5}
\vspace{-0.225in}
\end {table*}

\subsection{Interpretability Evaluation }
Any method for enhancing the interpretability of black-box models has to be rigorously evaluated. Measuring interpretability is not a straightforward process as there is no agreed-upon definition of interpretability in machine learning yet~\citep{doshi2017towards}. We base our evaluation of interpretability on the work of ~\cite{yang2019evaluating}, considering three criteria: generalizability, fidelity, and persuasibility. Also, we compare the interpretability of our model with two conventional machine learning models, i.e., support vector machine (SVM) and logistic regression (LR), as the baseline methods. 
\subsubsection{Evaluation On Generalizability}
The proposed interpretability method in this paper is based on the intrinsic characteristics of the activation functions used in each individual deep model (either ReLU or tanh). These activation functions either show locally linear or approximately linear behavior. Thus we consider this model to be semi-intrinsically interpretable. ~\cite{yang2019evaluating} define the generalizability evaluation of intrinsic interpretability task to be equivalent to the model evaluation using performance evaluation metrics such as accuracy, precision, and recall. According to the aforementioned results, the proposed consensus-based model significantly outperforms shallow models as well as the 5 individual deep-learning-based models.

\subsubsection{Evaluation On Fidelity}
In a previous section, we demonstrate how the proposed consensus-based algorithm can effectively reject irrelevant samples. To further examine this capability of the proposed algorithm, similar experiments are conducted on the MNIST image dataset. The proposed algorithm can successfully reject adversarial and irrelevant samples. 
%Further details are provided in the Supplementary Materials.
Further details can be found in the previous version of the work.

\subsubsection{Evaluation On Persuasibility}
A cardiologist, which is a co-author of this paper evaluates the validity of the interpretability method in the real-world setting. These evaluations show that if a patient has issues with other organ systems, he/she is in higher risk for developing a positive outcome ("die within a year") with the exception of infection and endocrinology.  In general, this indicates that a patient with issues with other organs is more susceptible to complications (i.e., comorbidities) along with AMI. Anemia diagnosed by hematocrit seems to significantly increase the risk of one-year mortality (positive outcome). Anemia defined by hemoglobin seems to be weakly predictive of one-year mortality. Liver and kidney dysfunction seem to be indicative of significant increased risk of one-year mortality which is consistent with the fact that the patient is generally sicker and has more critical conditions. Also, a cluster of procedures performed on patients decreases the risk of one-year mortality. This suggests that invasive procedures can decrease such a risk. In addition, two observations are noted. The first observation is about the procedures' naming in the data set. The lack of sufficient granularity of medical controlled vocabularies for coding procedure makes it difficult to interpret the importance of these procedures in the interpretability results. Also, some of the features that are strongly indicative of higher risk of one-year mortality seem to have an average within the normal range across the population. However, a closer look at their distribution profile in each class suggests that a slight deviation from the normal range associated with these features can enhance the risk for one-year mortality.

%%___As an instance, most of the categorical features related to undertaken treatment procedures are listed with negative impact on the positive class. In other words, these features have positive impact on the negative class (``did not die within a year"), which makes sense that more treatments should lead to a better outcome. Another example is the marital status, which is shown to act as neutral in the classification outcome, while cancer positive is shown to positively impact the positive class (``died within a year"). Patient age at the time of admission along with abnormal laboratory test values contributes positively to the positive class. The identified features are largely consistent with the features identified by other studies~\cite{yang2019predictors}. Also, those features with positive impact on the positive class are used in a conventional ICU mortality prediction tool APACHE-II~\cite{mercado2010apache}.__%%

\subsubsection{Comparisons with Baseline Evaluations}
Linear SVM and LR are considered to have intrinsic interpretability ~\citep{friedler2019assessing} and are quite popular for health data analysis~\citep{wang2018identification}. We also try the clusters of SVM (CSVM) by~\cite{gu2013clustered} to check if this ensemble method improves the performance of the shallow learner for this particular classification task. We set the number of clusters to 10 and observe that increasing it does not significantly enhance the model performance. Table~\ref{table:4} includes a comparison of the performance of LR, SVM, CSVM, and the proposed consensus-based algorithm.
To compare the feature-importance list as a result of interpretability enhancement of the consensus-based model to that of baseline models, we group features into top-n-features from n=10 to n=100 with step-size=10 and then calculate the percentage of similarity ($\sharp$ features ranked with same priority by both models/n).
% For example, the comparison on 50 features means we compare top-25 features contributing positively to the positive class and top-25 features contributing negatively to the positive class from the consensus-based algorithm to the same from the baseline models. 
The detailed comparison is provided in Table~\ref{table:5}. On average, the proposed consensus-based model shows 0.73 agreement with LR and 0.74 agreement with SVM on the feature-importance. These results confirm the fact that the proposed model shows linear behavior in local regions. 
%-- need to mention the intrinsic interpretability of our model 

\section{Related Work}
The lack of interpretability of deep neural networks is a limiting factor
of their adoption by healthcare and clinical practices.
Improving interpretability while retaining high accuracy of
deep neural networks is inherently challenging.
Existing interpretability enhancement methods can be categorized into integrated 
%(intrinsic interpretation) 
and post-hoc 
%(extrinsic interpretation)
approaches~\citep{adadi2018peeking}. The integrated methods utilize intrinsically interpretable models~\citep{melis2018towards} but they usually suffer from lower performance compared to deep models.
%as they can not approximate complex decisions in deep neural networks well. 
%Generally, higher prediction accuracy is achieved by more complex models~\cite{breiman2001statistical}. 
In contrast, the post-hoc interpretation methods attempt to provide explanations on an uninterpretable black-box model~\citep{koh2017understanding}. 
Such techniques can be further categorized into local and global interpretation ones. The local interpretation methods determine the importance of specific features to the overall performance of the model. The interpretable models are those that can explain why the system results in a specific prediction. This is different from the global interpretability approach, which provides a certain level of transparency about the whole model~\citep{du2018techniques}.
%For example back-propagation methods calculate the gradient of a
%particular output with respect to the input. The larger the magnitude of the gradient is, the stronger relevance of feature to prediction it demonstrates.
%% Salman here you need to compare your interpretable model to any work that is targetted for this comparison , I gave my own idea
The proposed method has the advantages of both local and global ones.
Our method relies on the local Jacobian difference vector to capture the importance of input features. At the same time, clusters of the difference vectors capture robust model behavior supported by multiple
training samples, reducing the complexity while retaining high accuracy.

\section{Conclusion and Future Work}

In this paper, we have proposed an interpretability method by
clustering local linear models of multiple models, capturing
feature importance compactly using cluster means.
Using consensus of multiple models allows us to improve 
classification accuracy and
interpretation robustness. Furthermore, the proposed deep ($n$, $k$)
consensus algorithm overcomes overgeneralization to irrelevant inputs
and oversensitivity to adversarial examples, which is necessary to be able to have meaningful interpretations.
For critical applications such as health care, it would be essential if
causal relationships between features and the outcomes can be identified and 
verified using existing medical knowledge. This is being further investigated.
%Our results seem to resolve the deep neural network paradox, where
%models with many parameters generalize well, which we will investigate
%further systematically.

%\newpage
\bibliographystyle{aaai}
\bibliography{aaai20}
\end{document}